# The Forgotten Code: Validating a Century-Old Translation System with AI

Jean-Marie Le Ray
Independent researcher, Rome, Italy
jmleray@translation2.com

**Abstract—***A pioneering rule-based mechanical translation system (precursor of modern RBMTs) was first presented in December 1929 by its inventor, Federico Pucci, who later published the full method in a book titled* "Il traduttore meccanico ed il metodo per corrispondersi fra Europei conoscendo ciascuno solo la propria lingua: Parte I"*, in Salerno (Italy), in 1931. This study illustrates how AI breathes new life into the system of international keys and ideograms devised by Pucci to translate from/into any Romance language (at least as a first step). The methodology involves having the AIs retranslate, following Pucci's method, the two text excerpts originally translated in 1931 and clearly documented in his publication: a passage from Dante's La Vita Nuova, translated from Italian into French, and a passage from Voltaire's Zadig, translated from French into Italian. The result is notable: the two texts, translated 94 years apart using the same method — by Pucci in 1931 and by AIs in 2025 — show a low average difference, with only minor variations observed. With Pucci's system thus validated, it became feasible to have the AIs reproduce the excerpts in English, Spanish, and German according to his method. The results were consistent, and Pucci — via Artificial Intelligence — was tasked with translating more modern and technical texts, thereby reviving, nearly a century later, an invention that had remained almost entirely unknown and never applied beyond its creator, now brought to wider attention and opened to possible experimentation. Such a demonstration would not only affirm Pucci's historical status but also place him among the precursors and intellectual contributors to machine translation, whose work merits examination alongside figures such as Troyanskij, Booth, and Weaver, with possible consequences for how the history of the field is understood.*

## 1. Introduction

Throughout history, the phenomenon of simultaneous invention—where multiple individuals independently arrive at the same discovery around the same time—has highlighted how scientific and technological breakthroughs often emerge from shared intellectual, social, and technological contexts rather than isolated genius. Great minds think alike! One of the most illustrative cases is the decipherment of Egyptian hieroglyphs. In the early 19th century, Jean-François Champollion in France and Thomas Young in the United Kingdom both worked on interpreting the Rosetta Stone, a trilingual artifact discovered in 1799 during Napoleon's Egyptian campaign. Their separate yet converging efforts, facilitated by access to similar resources and a growing knowledge of Coptic, illustrate how a ripe intellectual environment can catalyze multiple independent discoveries. This pattern extends well beyond ancient scripts. In fact, the development of machine translation (MT) provides another compelling instance of simultaneous invention. Before the advent of modern computers, thinkers in different countries were already imagining how machines might bridge linguistic barriers. During the 1930s and 1940s, ideas about automating translation were forming in parallel across Europe and the Soviet Union. These early efforts responded to a growing need for multilingual communication, especially during and after the disruptions of World War II:

*Then the war came and I attempted to steer my studies towards military use. I managed to create mechanical translating devices "C" and "D", a mechanical solution, attempting to create a new mechanical-based language, with device C working as a transmitter, and D as a receiver device.* ***They were to be submitted to the 1940 Engineering Exhibition, but the War Ministry opposed its participation. I was called to Rome to explain the invention.*** *It was approved, and I was authorized to build and try out the device, at the State's expense, since I had informed them, I could not afford to build it on my own. Obviously, I was obliged to keep everything secret. However, as I was not a mechanic, I thought that I would need the assistance of other people, who might not be able to keep the secret. I did not want to run this risk, so I turned down the assignment, and left the invention in the hands of the War Ministry, so that it might do whatever it wanted with the idea.*

These words, written by Federico Pucci in his first letter to the CNR, the Italian National Research Council, on July 10, 1949, sought to claim his anteriority for a "translation machine" he had first envisioned twenty years earlier (see Appendix A, Figure A.1). His method, first laid out in detail in a 1931 book published in Salerno, marks a seminal moment in the early prehistory of machine translation. The work, titled *Il traduttore meccanico ed il metodo per corrispondersi fra europei, conoscendo ciascuno solo la propria lingua: Parte I* [4] (see Appendix B, Figure B.1), offers what can be considered the earliest blueprint of a Rule-Based Machine Translation (RBMT) system.

This early publication also includes two translated passages, one from Dante's La Vita Nuova and another from Voltaire's Zadig, rendered respectively from Italian into French and French into Italian. These represent perhaps the first known examples of "machine translation," even if the term itself would not be coined until Warren Weaver's influential 1949 memorandum Translation [5] and later popularized in 1952 at the first international conference on MT at MIT [6].

In the preface dated December 10, 1930, Pucci explained that his method aimed at enabling communication among people who spoke different European languages, each relying only on their native tongue. His proposal, however, remained theoretical. There is no record of a functioning prototype, nor any institutional support for turning his vision into a reality.

Pucci's efforts were paralleled, unknown to him, by other inventors. In 1933, Georges B. Artsrouni in France and Petr Petrovich Troyanskii in the Soviet Union filed patents for mechanical translation devices [8]. Artsrouni's patent, granted on July 22, 1933, described a "mechanical brain," a general-purpose machine that could also function as a multilingual dictionary. He had begun work as early as 1929 and completed a prototype by 1932. While the machine was demonstrated in Paris in 1937 and won accolades, it essentially performed lexical lookups via punched tape and did not produce full translations.

Troyanskii, on the other hand, filed a Soviet patent on September 5, 1933, which was granted in 1935. His device was misunderstood by Soviet authorities as a typesetting tool, delaying its recognition. Only in 1959 did the Soviet Academy of Sciences bring his work to broader attention. Like Pucci and unlike Artsrouni, Troyanskii envisioned a complete translation system. His design featured grammatical coding, a symbolic interlingua inspired by Esperanto, and a three-phase process for converting source texts into target languages. His theoretical model laid the groundwork for future developments, although, like Pucci's, it was never physically constructed.

As the MT researcher John Hutchins noted, Troyanskii's engineering focus contrasted with his limited linguistic grounding [14]. The inverse might be said of Pucci: a linguist by training, his grasp of mechanical implementation was minimal. Yet in conceptual scope, Pucci's work aligns more closely with Troyanskii's than with Artsrouni's, particularly in addressing syntactic and morphological structures rather than mere word matching.

Despite their differences, all three pioneers were part of a wider movement responding to the intellectual climate of the early 20th century—a time when global communication was becoming increasingly urgent. Still, Pucci's foresight was particularly remarkable. His work predates the Georgetown-IBM experiment of 1954 [15], often cited as the field's foundational demonstration, by a full 25 years. And unlike Warren Weaver's 1949 proposal 20 years later, which assumed the use of computers for translation [11], Pucci formulated his ideas before digital technologies were even feasible.

It is striking, then, that Pucci has remained virtually unknown. During his life, his work received little attention. In death, it was forgotten entirely. Only in March 2017 was his 1931 publication rediscovered. An original copy is preserved in the National Central Library of Florence, but no scholar had previously cited or studied it. Its rediscovery was prompted by brief mentions in early writings by John Hutchins, references that have since vanished from the web (see Appendix C, Figure C.1). In the first one [9], Hutchins said:

*In August 1949, the New York Times reported from Salerno that an Italian named Federico Pucci, had invented a machine to translate, saying that it would be exhibited at a Paris Fair;* ***but no more was to be heard of it****.*

Then in a second updated version [10, 16]:

*(…) The second consists of an intriguing announcement from Italy just three months after the newspaper report of SWAC. On 26 August 1949, the New York Times reported (page 9) from Salerno:*

*Federico Pucci announced today that he had invented a machine that could translate copy from any language into any other language. He said that the machine was electrically operated, but refused to disclose details. He said that he would enter it in the Paris International Fair of Inventions next month.*

*It is uncertain whether Pucci had any knowledge of Huskey's proposals, and it seems most unlikely he knew about Weaver's memorandum or the British experiments. In any event, there is no trace of any demonstration at the Paris fair;* ***and nothing more is known about Pucci****.*

This lack of recognition is particularly glaring considering the trilingual documentation available and the profound alignment of Pucci's concepts with modern translation paradigms. Pucci envisioned communication without multilingual fluency, relying instead on a structured system of linguistic equivalences—a principle that now underpins much of AI-driven language processing.

As we approach the centenary of Pucci's first public presentation in December 1929, the time is ripe to reexamine his contribution. This article argues that Pucci should be acknowledged as a pioneering figure in the pre-digital history of machine translation. His conceptual work anticipated many of the field's later developments, including rule-based approaches, "plain language" and "good enough" concepts (see Appendix D) and symbolic reasoning. With the capabilities of today's AI systems to handle morphological parsing, interlingual representation and semantic alignment, we are now able to implement Pucci's method.

In fact, Pucci devised a kind of mechanical Rosetta Stone for the 21st century, reflecting the same multilingual logic that made the original stele such a revolutionary tool which enabled scholars to decode a lost script. That his method was not realized in hardware is less significant than the originality of its conceptual architecture. Pucci's vision was robust enough to survive decades of technological evolution and could have served as a theoretical foundation for mid-20th-century efforts—had anyone known of it.

This study aims to correct that historical oversight. It seeks not only to document Pucci's forgotten contributions but also to reassess his rightful place among the early thinkers in machine translation. Had his method informed the Georgetown-IBM experiment or Warren Weaver's foundational theories, the trajectory of the field might have been significantly different.

Moreover, Pucci's example challenges us to reconsider the criteria by which we define innovation. Is it the production of a working prototype, or the articulation of a new conceptual framework? In a field where early prototypes were often crude and impractical, Pucci's theoretical sophistication deserves renewed attention.

Ultimately, the rediscovery of Pucci's work opens a new chapter in the historiography of machine translation. It invites scholars, engineers, and linguists alike to revisit the field's origins and to consider that the seeds of today's language technologies were planted not only in postwar America or the Soviet Union, but also in interwar Italy by a little-known linguist with a visionary mind.

*Research hypotheses and questions.* From the research, it is possible to derive a rich set of hypotheses and questions at the intersection of machine translation, AI historiography, philosophy of technology and epistemic justice. Hypotheses are the following:

- Pucci’s method, though conceived in 1931, is sufficiently rigorous and structured to be operationalized by contemporary AI systems for accurate translations between Romance languages.
- Pucci’s work remained obscure not because of its lack of merit, but due to contextual limitations (technological, geographic, political, or linguistic) and historiographical bias.
- The reactivation of Pucci’s system through AI constitutes both a continuation and a transformation of the original concept, challenging traditional ideas of technological identity.
- Artificial intelligence can serve as an active agent of historical revisionism by enabling the realization and testing of forgotten or unbuilt systems, thereby rewriting historical trajectories.
- Bringing Pucci’s system to light through AI constitutes an act of epistemic justice, restoring a voice and value to a previously neglected contributor in the field of machine translation.

Questions are:

- To what extent can modern AI accurately reproduce translations using Federico Pucci’s 1931 system of mechanical translation? How robust is Pucci’s rule-based translation method when applied to modern and technical texts using AI?
- What does the successful implementation of Pucci’s method by AI suggest about his role in the early history of MT? And how does this case disrupt the linear narrative of technological progress in the field?
- Is a technological system the same if it is revived decades later in a different material form (AI vs. mechanical)? Can artificial intelligence function as a medium of historical memory and technological resurrection?
- What constitutes the identity of a machine translation system — its physical realization, conceptual design, or function? How do AI-driven revivals challenge our definitions of technological “invention” and “authorship”?
- Can AI help restore forgotten or marginalized contributions to scientific and technological history? What mechanisms contribute to the erasure or omission of such contributions in dominant narratives?

*Article structure.* The research presented in this paper adopts a multidisciplinary approach, incorporating the fields of machine translation and language technologies, as well as practical applications of generative AI models. Section 2 outlines the methodology, sources, reconstruction of Pucci's method, and the AI systems used. Section 3 addresses possible objections and presents a detailed AIs translation process according to the Pucci's method, along with a control test between quantitative results produced by AIs and between those and Pucci's output. Section 4 examines the findings, including a qualitative assessment of the data and their interpretation, limitations, and ambiguities—what these results suggest about the robustness of Pucci's method, and how it can be applied with AI to translate more modern and technical texts in other languages. Section 5 formulates recommendations aimed at reconfiguring the historiography of machine translation, situating Pucci within a revised narrative that acknowledges the broader implications of the present study for a reassessment of his pioneering role. Section 6 concludes.

## 2. Methodology

### 2.1 Source material

To the best of current knowledge, Federico Pucci authored ten works dedicated to his invention over a period spanning from 1931 to 1960 (see Appendix E, Figure E.1). This study draws upon these primary sources, which are preserved in their entirety in various Italian libraries, most notably in Florence and Salerno. Among these publications, the most significant is undoubtedly the inaugural volume: *Il Traduttore Meccanico ed il Metodo per corrispondersi fra Europei conoscendo ciascuno solo la propria lingua: Parte I* (1931) [4], a foundational text. An original copy of this work is held in the *Biblioteca Nazionale Centrale di Firenze* (see Appendix B, Figure B.1). The second most important text is the seventh, published in 1950, known colloquially as the "book-machine." (see Appendix F, Figure F.1). This volume includes a functional prototype—a model designed to serve as the basis for the construction of Pucci's envisioned translation machine—along with a detailed explanation of its operational logic (see Appendix F, Figure F.3). Within this framework, section A designates the mobile vocabulary, B represents its complementary extension, C functions as a syntactic corrector, and D as a morphological corrector (see Appendix F, Figure F.4). At least two extant copies of this book are known: one housed in Salerno (see Appendix F, Figure F.2) and another in Florence, each containing an identical hand-crafted model assembled by Pucci himself. While the exact print run of this volume is unknown, it undoubtedly represents an enormous amount of work and considerable personal investment. In his own writings, Pucci claimed that his machines could take multiple forms—including simple mechanical, electrical, phono-electrical, photo-electrical, and tele-electrical devices—and that they could give rise to various complex derivatives. Among these was the *Interprète Électro-mécanique Portable*, which received a silver medal at the 1950 International Exposition in Liège, as part of the Grand Invention Competition. The remaining volumes primarily consist of bilingual grammars and specialized vocabularies (mainly in Italian, French, English, and German), intended to facilitate the practical application of Pucci's translation system.

The reconstruction of Pucci's method was carried out by faithfully reproducing his system according to his own instructions. This involved digitally encoding his international key tables and the ideographic symbols he devised and subsequently submitting this structured data to a range of generative AI systems for analysis and implementation, prompting essentially Figures 1 & 2 (Charts "A", "B" and basic keys according to Pucci) and submitting this structured data to generative AI systems with a simple prompt: '*Translate the Dante excerpt using Pucci's method.*' To comply, each AI had to implicitly rebuild Pucci's system from the charts provided—a critical point that demonstrates the remarkable capacity of modern LLMs to infer complex, rule-based systems from minimal data. This is the focus of the next section.

## 2.2 Reconstruction of Pucci's method

Back to the early 1930s, Federico Pucci illustrated his method for "automatic" translation without requiring the reader to know the other language. His stated aim was to allow individuals, including those with limited education, to comprehend foreign speech rendered into their own language.

Even by today's standards, translating some of Pucci's demonstration texts would be a complex task. Therefore, it is difficult to overstate the visionary nature of Pucci's proposal: he aimed to achieve mechanical translation nearly a century ago, in an era long before digital computation, and promised that one could learn to translate in under a minute. In 1950, reflecting his non-commercial intent, Pucci priced his publication at just 150 lire for the book, and 600 lire (approximately 10 U.S. dollars at the time) for the version including the machine model (see Appendix F, Figure F.1). The envisioned machine, which he hoped would be both portable and economical, never moved beyond the blueprint stage. Nonetheless, the conceptual leap he made placed him nearly a century ahead of his time.

The core of Pucci's invention rested on three principal steps: first, breaking down sentences into their smallest semantic units (morphemes); second, transposing those units into a target language using a system of ideographic codes; and third, allowing the native-speaking recipient to reorder the translated elements into correct and fluent syntax.

This approach, quite ambitious, is very close to the three-step system proposed later by Troyanskii: a grammatical encoding based on Esperanto, with human pre-editing, mechanical translation, and post-editing, which was far ahead of a modern concept of machine translation.

In doing so, Pucci also anticipated two ideas that are foundational in contemporary MTPE: language simplification as the input — a practice that may be viewed as a precursor to the notion of *Plain Language* — and a translation characterized by the "*good enough*" concept as the output (see Appendix D). The most pivotal elements of his system, logical and functional, was supported by ideograms—both basic and derived—contained in two critical charts of "international keys" and a set of operational rules. These form the heart of his invention and constitute the symbolic code that his proposed machine would use to mechanically translate texts.

The chart "A" defines grammatical functions using ideograms applicable to Romance languages, and —with slight modifications— also to Germanic and Slavic languages. This chart includes symbols for articles, personal pronouns, possessives, demonstratives, and relative pronouns. For example, "m" and "f" indicate masculine and feminine, while "I" to "III+" denote pronouns from first to third person, singular and plural.

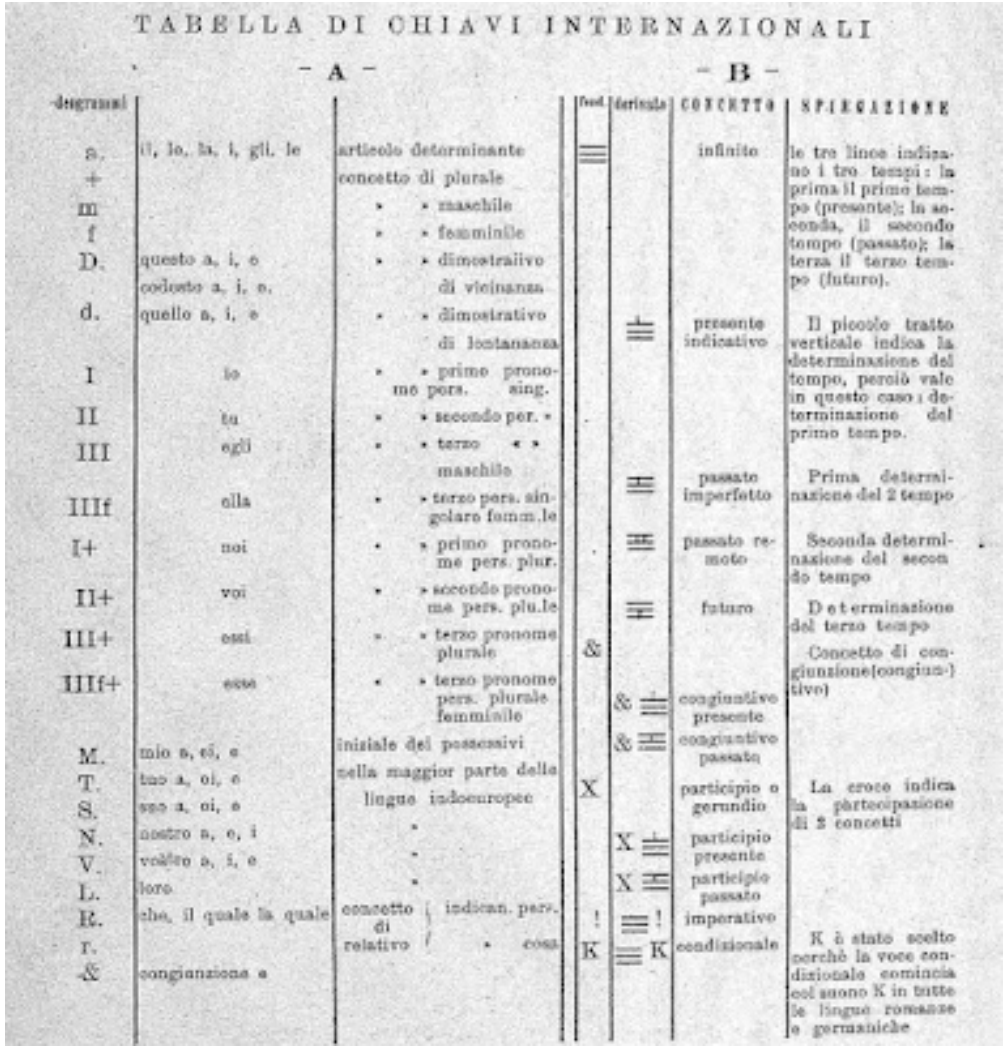
TABELLA DI CHIAVI INTERNAZIONALI

– A –

| ideogrammi | | |
|---|---|---|
| a. | il, lo, la, i, gli, le | articolo determinante |
| + | | concetto di plurale |
| m | | » » maschile |
| f | | » » femminile |
| D. | questo a, i, e / codesto a, i, e, | » » dimostrativo di vicinanza |
| d. | quello a, i, e | » » dimostrativo di lontananza |
| I | io | » » primo pronome pers. sing. |
| II | tu | » » secondo per. » |
| III | egli | » » terzo » » maschile |
| IIIf | ella | » » terzo pers. singolare femm.le |
| I+ | noi | » » primo pronome pers. plur. |
| II+ | voi | » » secondo pronome pers. plu.le |
| III+ | essi | » » terzo pronome plurale |
| IIIf+ | esse | » » terzo pronome pers. plurale femminile |
| M. | mio a, ei, e | iniziale dei possessivi nella maggior parte delle lingue indoeuropee |
| T. | tuo a, oi, e | |
| S. | suo a, oi, e | |
| N. | nostro a, e, i | » |
| V. | vostro a, i, e | » |
| L. | loro | » |
| R. | che, il quale la quale | concetto di relativo – indican. pers. |
| r. | | concetto di relativo – » cosa |
| & | congiunzione e | |

– B –

| fond. | derivate | CONCETTO | SPIEGAZIONE |
|---|---|---|---|
| ≡ | | infinito | le tre linee indicano i tre tempi: la prima il primo tempo (presente); la seconda, il secondo tempo (passato); la terza il terzo tempo (futuro). |
| | [ideogram] | presente indicativo | Il piccolo tratto verticale indica la determinazione del tempo, perciò vale in questo caso: determinazione del primo tempo. |
| | [ideogram] | passato imperfetto | Prima determinazione del 2 tempo |
| | [ideogram] | passato remoto | Seconda determinazione del secondo tempo |
| | [ideogram] | futuro | Determinazione del terzo tempo |
| & | | | Concetto di congiunzione (congiuntivo) |
| | & [ideogram] | congiuntivo presente | |
| | & [ideogram] | congiuntivo passato | |
| X | | participio o gerundio | La croce indica la partecipazione di 2 concetti |
| | X [ideogram] | participio presente | |
| | X [ideogram] | participio passato | |
| ! | ≡ ! | imperativo | |
| K | ≡ K | condizionale | K è stato scelto perchè la voce condizionale comincia col suono K in tutte le lingue romanze e germaniche |

**Figure 1**
Charts "**A**" and "**B**" of Pucci's international keys and ideograms

The chart “B” introduces derived forms for verbal tense and mood. It includes markings for infinitive, present indicative, imperfect, remote past, future, subjunctive, participle, imperative, and conditional. Each form is represented by a logical, abstract ideogram —for instance, the letter “K” for the conditional, referencing the initial phoneme common to many Indo-European languages.

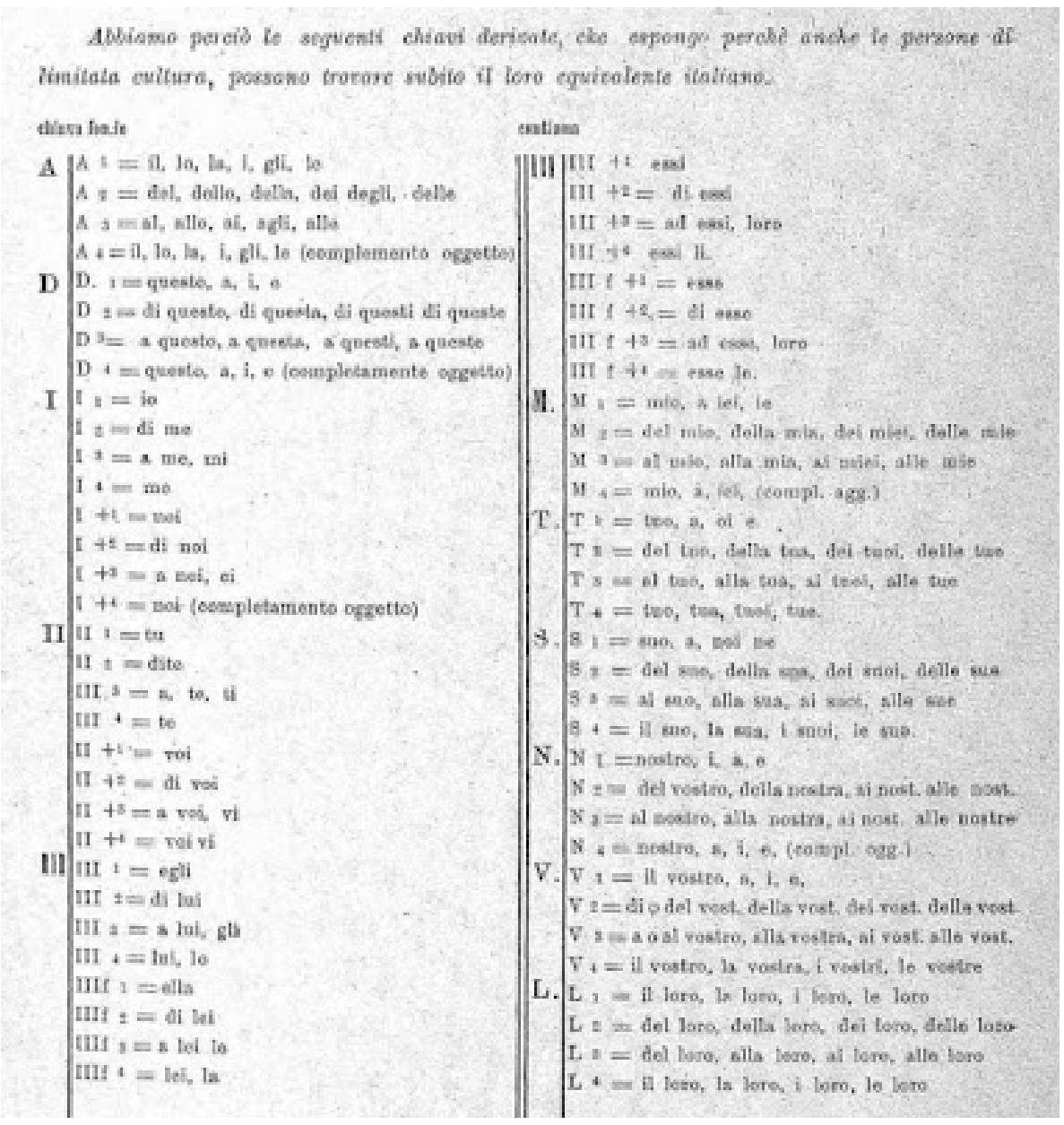
*Abbiamo perciò le seguenti chiavi derivate, che espongo perchè anche le persone di limitata cultura, possono trovare subito il loro equivalente italiano.*

chiave base | continua

A
A 1 = il, lo, la, i, gli, le
A 2 = del, dello, della, dei degli, delle
A 3 = al, allo, ai, agli, alle
A 4 = il, lo, la, i, gli, le (complemento oggetto)

D
D. 1 = questo, a, i, e
D 2 = di questo, di questa, di questi di queste
D 3 = a questo, a questa, a questi, a queste
D 4 = questo, a, i, e (complemento oggetto)

I
I 1 = io
I 2 = di me
I 3 = a me, mi
I 4 = me
I +1 = noi
I +2 = di noi
I +3 = a noi, ci
I +4 = noi (complemento oggetto)

II
II 1 = tu
II 2 = di te
II 3 = a, te, ti
II 4 = te
II +1 = voi
II +2 = di voi
II +3 = a voi, vi
II +4 = voi vi

III
III 1 = egli
III 2 = di lui
III 3 = a lui, gli
III 4 = lui, lo
IIIf 1 = ella
IIIf 2 = di lei
IIIf 3 = a lei le
IIIf 4 = lei, la

III
III +1 essi
III +2 = di essi
III +3 = ad essi, loro
III +4 essi li.
III f +1 = esse
III f +2 = di esse
III f +3 = ad esse, loro
III f +4 = esse le.

M.
M 1 = mio, a, iei, ie
M 2 = del mio, della mia, dei miei, delle mie
M 3 = al mio, alla mia, ai miei, alle mie
M 4 = mio, a, iei, (compl. agg.)

T.
T 1 = tuo, a, oi e
T 2 = del tuo, della tua, dei tuoi, delle tue
T 3 = al tuo, alla tua, ai tuoi, alle tue
T 4 = tuo, tua, tuoi, tue.

S.
S 1 = suo, a, uoi ue
S 2 = del suo, della sua, dei suoi, delle sue
S 3 = al suo, alla sua, ai suoi, alle sue
S 4 = il suo, la sua, i suoi, le sue.

N.
N 1 = nostro, i, a, e
N 2 = del nostro, della nostra, ai nost. alle nost.
N 3 = al nostro, alla nostra, ai nost. alle nostre
N 4 = nostro, a, i, e, (compl. ogg.)

V.
V 1 = il vostro, a, i, e,
V 2 = di o del vost. della vost. dei vost. delle vost.
V 3 = a o al vostro, alla vostra, ai vost. alle vost.
V 4 = il vostro, la vostra, i vostri, le vostre

L.
L 1 = il loro, la loro, i loro, le loro
L 2 = del loro, della loro, dei loro, delle loro
L 3 = del loro, alla loro, ai loro, alle loro
L 4 = il loro, la loro, i loro, le loro

**Figure 2**
Second chart of basic keys by inflecting personal pronouns, articles, possessives, demonstratives, etc.

Pucci also outlined precise rules for using the system. For instance, certain words should be replaced by ideograms, with gender indicators attached to relevant adjectives (e.g., buon-m libro for "good [masculine] book"). Verbal forms such as porterebbe (he/she would bring) are broken down into their infinitive root and the conditional ideogram. Simplified expressions and lexical substitutions are encouraged, especially for idioms or rare phrases. The system even provides mechanisms to adapt gender and number between source and target languages by referencing dictionary-based correspondences.

## 2.3 Selection of AIs

Once Pucci’s key tables and ideographic symbols were digitally encoded, this structured data was submitted to various generative AI systems for analysis and implementation. The three AIs selected are versions freely available online: **GPT-4o**, **Claude Sonnet 4** and **Grok 3.** We did not choose Google Translate or DeepL because they are not trainable LLMs.

The strategic choice to exclusively utilize free AI tools was driven by a deliberately conservative methodology rooted in the concept of establishing a quality baseline. This approach operates on a fundamental premise: if satisfactory results can be achieved using "simplified" or basic AI systems, then the outcomes would undoubtedly be significantly superior when employing more sophisticated, premium AI versions. Furthermore, this baseline approach demonstrates what individuals, small organizations, or researchers with limited budgets can realistically achieve, while simultaneously indicating the potential for dramatic improvement when resources allow for more sophisticated implementations.

### 3. Detailed AIs translation process according to the Pucci's method

I must first address two objections regarding the translation of these texts—the first rendered by Pucci in 1931, and the second by AIs in 2025 following Pucci's method.

1. The first one might be that Mr. Pucci could very well have translated this passage himself without using his method.
2. The second objection was phrased as follows: "*I'm just concerned that those familiar with generative AI might criticize the approach, given the tendency of AI to hallucinate.*"

Indeed, these two concerns are closely connected, as they both revolve around the challenge of validating Pucci's method through AI, while simultaneously assessing the reliability of AIs based on how closely their outputs align with Pucci's original translations and source texts.

To address the first one, we must recognize Pucci's intellectual rigor and integrity. After more than eight years of in-depth research into his life and work, I am convinced that he was too serious a scholar to fake the results by translating conventionally and merely attributing it to his method. As a trained linguist, Pucci would not have expressed himself in certain stylistic ways found in the translation—precisely because they reflect a mechanical, systematic procedure rather than the choices of a human translator.

As for the second objection—the risk of AI hallucination—it is a valid point, especially given current discussions around generative models. However, this risk can be minimized through controlled experimentation.

So, to conduct this experiment rigorously, one could first digitize the original text translated by Pucci in 1931 and reconstruct his method as faithfully as possible based on his own descriptions. Then, using different large language models or AI translation systems—each instructed to apply Pucci's procedure step by step—researchers would generate independent translations of the same passage. The resulting outputs could then be aligned and analyzed both quantitatively (e.g., similarity scores, lexical overlap, syntactic structure) and qualitatively (e.g., stylistic features, error patterns). If the AI-produced translations converge significantly with Pucci's version, and do so consistently across different models, it would offer a strong support for both the internal coherence of Pucci's method and its forward-looking compatibility with contemporary machine translation frameworks.

Moreover, hallucination implies fabrication or invention beyond the input or instructions given. In this case, if one AI, under controlled application of Pucci's method, reproduces a translation nearly identical to what Pucci himself produced manually in 1931, it strengthens — not weakens — the argument for his prescient insight. AI becomes, in fact, a tool to simulate his reasoning, not an independent agent producing its own invention.

So, the objection underscores the value of rigorously replicating Pucci's process. If an AI hallucinated its way into matching a human translation from 1931 by chance, that would indeed be extraordinary — but highly improbable. The far more plausible explanation would be that Pucci's system was methodologically sound and anticipatory of principles that now govern machine translation. This collective consistency provides concrete proof, moving the conversation from 'curious coincidence' to 'replicable model,' demonstrating that Pucci's method anticipated key mechanisms of modern computational translation.

That is not a hallucination, but rather an act of empirical validation.

### 3.1 Comparison of encodings between Pucci and AIs

Of the two texts translated by Federico Pucci in 1931, this study confines its comparative analysis with contemporary AI-generated translations to the Dante excerpt, as the results are sufficient to demonstrate the methodological soundness and predictive accuracy of Pucci's system, while the translation of the second excerpt from French into Italian offers no additional comparative value. Here is the chosen passage of Dante's *La Vita Nuova*:

*Ai miei occhi apparve la gloriosa donna della mia mente, la quale fu da molti chiamata Beatrice. Io la vidi quasi dalla fine del mio anno nono. Apparve vestita di nobilissimo colore, cinta ed ornata alla guisa che alla sua giovanissima età si convenia. In quel punto dico veramente che lo spirito della vita cominciò a tremar sì fortemente che apparia nei menomi polsi orribilmente. E vedeala di sì nobili e laudabili portamenti, che si potea dire quella parola del poeta Omero: Ella non parea figliuola d'uomo mortale, ma di Dio.*
*Poi che furono passati tanti dì, nell'ultimo di questi, avvenne che questa mirabile donna apparve a me vestita di colore bianchissimo, in mezzo di due gentili donne, le quali erano di piu lunga età; e, passando per una via, volse gli occhi verso me e mi salutò molto virtuosamente, tanto che mi parve allora vedere tutti i termini della beatitudine.*

According to his own method, Pucci coded the passage to be translated as follows.

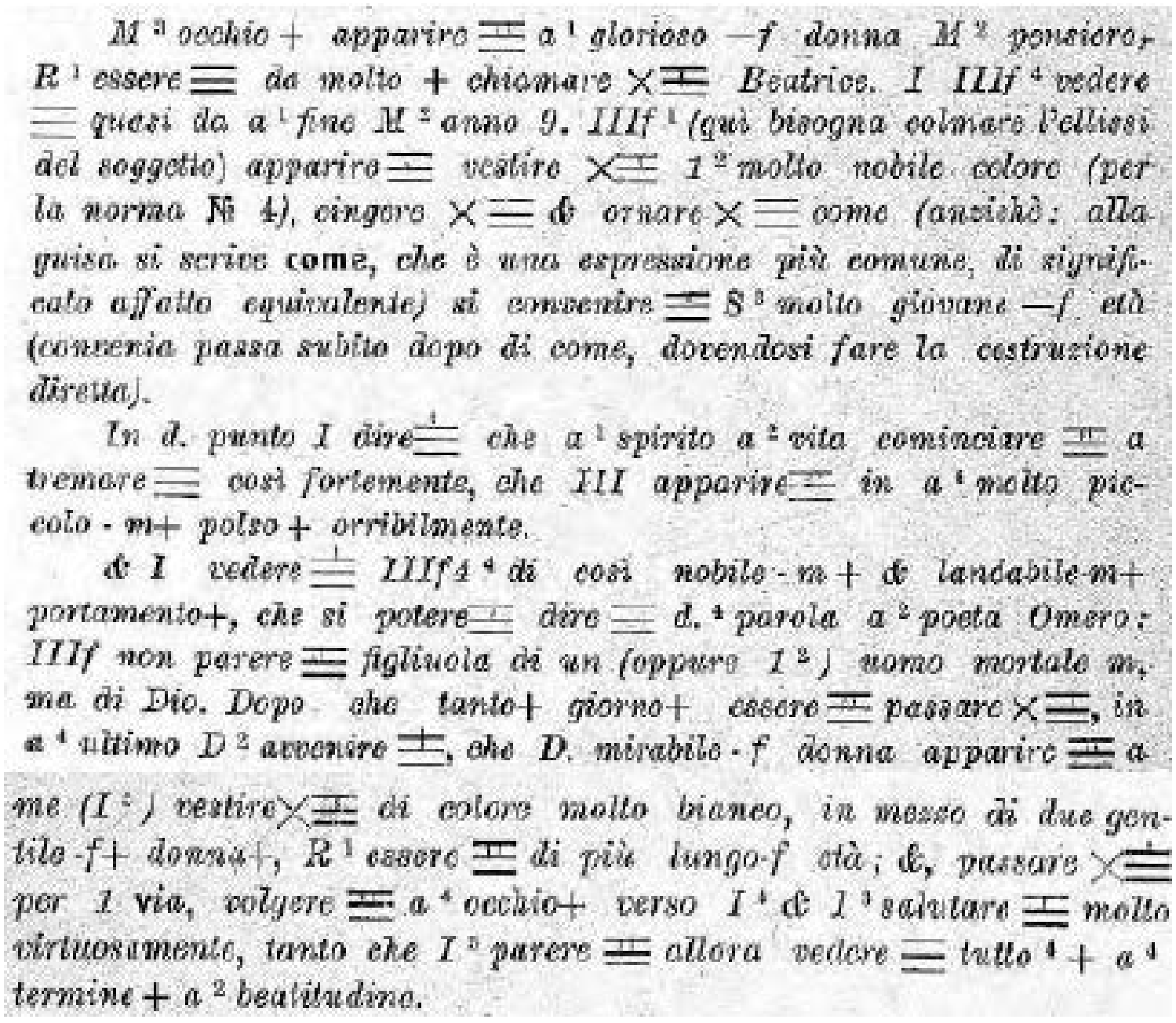
*M $^{2}$ occhio + apparire ≡ a $^{1}$ glorioso —f donna M $^{2}$ pensiero, R $^{1}$ essere ≡ da molto + chiamare ×≡ Beatrice. I IIIf $^{4}$ vedere ≡ quasi da a $^{1}$ fine M $^{2}$ anno 9. IIIf $^{1}$ (qui bisogna colmare l'ellissi del soggetto) apparire ≡ vestire ×≡ 1 $^{2}$ molto nobile colore (per la norma № 4), cingere ×≡ & ornare ×≡ come (anzichè: alla guisa si scrive* **come**, *che è una espressione più comune, di significato affatto equivalente) si convenire ≡ S $^{3}$ molto giovane —f età (convenia passa subito dopo di come, dovendosi fare la costruzione diretta).*

*In d. punto I dire ≡ che a $^{1}$ spirito a $^{2}$ vita cominciare ≡ a tremare ≡ così fortemente, che III apparire ≡ in a $^{4}$ molto piccolo - m+ polso + orribilmente.*

*& I vedere ≡ IIIf4 $^{4}$ di così nobile - m + & laudabile-m+ portamento+, che si potere ≡ dire ≡ d. $^{4}$ parola a $^{2}$ poeta Omero: IIIf non parere ≡ figliuola di un (oppure 1 $^{2}$) uomo mortale m, ma di Dio. Dopo che tanto+ giorno+ essere ≡ passare ×≡, in a $^{4}$ ultimo D $^{2}$ avvenire ≡, che D. mirabile - f donna apparire ≡ a me (I $^{2}$) vestire ×≡ di colore molto bianco, in mezzo di due gentile -f+ donna+, R $^{1}$ essere ≡ di più lungo-f età; &, passare ×≡ per 1 via, volgere ≡ a $^{4}$ occhio+ verso I $^{4}$ & I $^{3}$ salutare ≡ molto virtuosamente, tanto che I $^{3}$ parere ≡ allora vedere ≡ tutto $^{4}$ + a $^{4}$ termine + a $^{2}$ beatitudine.*

**Figure 3**
Encoding of original Italian text by Dante, from *La Vita Nuova* (page 27 of Pucci's book)

The translated Pucci's result into French is this one:

*À mes yeux apparut la glorieuse femme de ma pensée, laquelle était par bien des personnes appelée Béatrice. Je la vis depuis la fin de mon année neuvième. Elle apparut habillée d'une très noble couleur, ceinte et ornée comme il se convenait à son très jeune âge. À ce point je dis vraiment que l'esprit de la vie commença à trembler si fortement qu'il apparaissait dans les très petits pouls horriblement. Et je la voyais de si nobles et louables contenances qu'on pouvait dire cette parole du poète Homère: elle ne semblait pas fille d'un homme mortel, mais de Dieu.*

*Après que tant de jours furent passés, dans le dernier de ceux-ci, il arriva, que cette femme admirable apparut à moi, habillée d'une couleur très blanche au milieu de deux femmes de condition, qui étaient d'un plus long âge ; en passant elle tourna les yeux vers moi et me salua très vertueusement de sorte que il me parut alors de voir tous les limites de la béatitude.*

Regarding encoding the text by AIs, since I couldn't load the graphic ideograms saying LLMs their meaning, I gave them a text equivalent as follows: ID1, ID2, ..., ID11 to describe verbal tense:

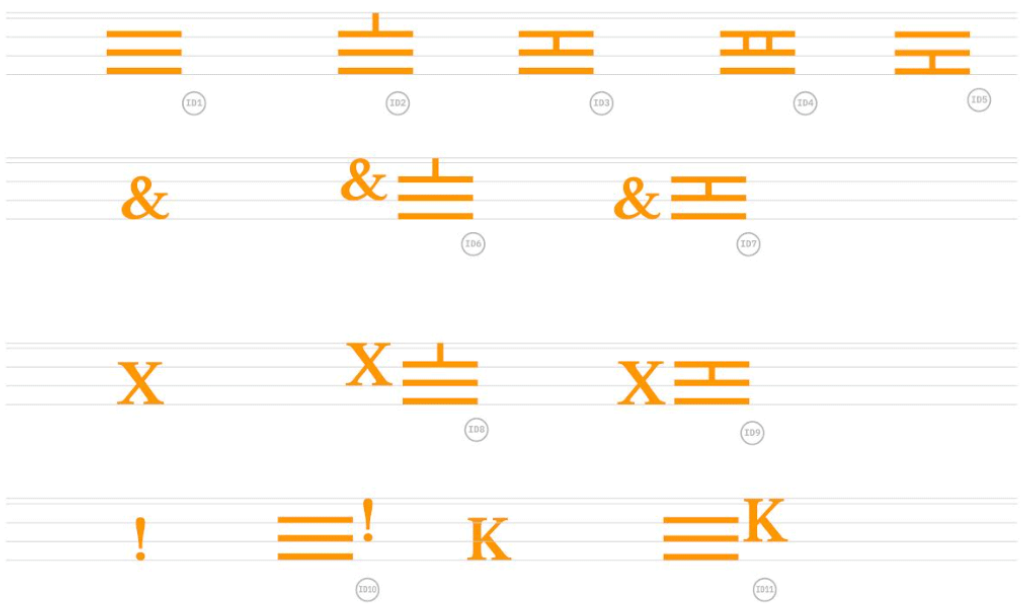

**Figure 4**
Numbering of graphic ideograms (except &, X, ! and K) from ID1 to ID11

Where ID1 = infinitive, ID2 = present indicative, ID3 = imperfect, ID4 = remote past, ID5 = future, ID6 = present subjunctive, ID7 = past subjunctive, ID8 = present participle, ID9 = past participle, ID10 = imperative and ID11 = conditional.

Encoding of original Italian text by ChatGPT:

a. M. yeux apparaître-ID4 a. glorieuse femme de M. esprit, r. être-ID4 appelée Beatrice par plusieurs. I la voir-ID4 presque à la fin de M. année neuvième. Apparaître-ID4 vêtue d'un très noble couleur, ceinte et ornée selon la manière convenable à S. jeune âge. En ce moment I dire-ID2 vraiment que l'esprit de la vie commencer-ID4 à trembler si fortement qu'il apparaître-ID3 horriblement dans les moindres pouls. Et I la voir-ID3 de si nobles et louables manières qu'on pouvoir-ID3 dire cette parole du poète Homère : Elle ne sembler-ID3 fille d'un homme mortel, mais de Dieu. Puisque plusieurs jours être-ID4 passés, au dernier de ceux-ci, il arriver-ID4 que D. admirable femme apparaître-ID4 à I, vêtue de couleur très blanche, entre deux femmes gentilles, r. être-ID3 de plus grande âge ; et, en passant par une rue, tourner-ID4 les yeux vers I et I saluer-ID4 très vertueusement, tellement que il I sembler-ID4 alors voir tous les confins de la béatitude.

**Figure 5**
Encoding of original Italian text by ChatGPT

Encoding of original Italian text by Claude (apparently more sophisticated than that of ChatGPT):

**First Paragraph:** À **M.** yeux **apparir-ID4 a.** glorios-f **donn-f** de **a. M.** ment-f, **a. r. être-ID4** de beaucoup **appeler-ID9** Béatrice. **I a. voir-ID4** presque de **a.** fin de **M. ann-f** neuvièm-f. **Apparir-ID4** vêtu-f de très-nobl-m **couleur-f**, ceint-f & **orner-ID9** à **a.** façon **r.** à **S.** très-jeun-f **âg-m se convenir-ID3**. Dans ce **point-m dir-ID2** vraiment **r. a. esprit-m** de **a. vi-f commencer-ID4** à **trembler** si fortement **r. apparir-ID3** dans **a.** moindre **poul-m** horriblement. & **a. voir-ID3** de si **nobl-m** & **louabl-m comportement-m**, **r. se pouvoir-ID3 dir** cette **parol-f** du **poèt-m** Homère: **IIIf** ne **paraîtr-ID3 fill-f** d'**homm-m mortel-m**, mais de Dieu.¶

**Second Paragraph:** Après **r. être-ID4 passer-ID9** tant **jour-m**, dans **a.** dernier de ces, **advenir-ID4 r.** cette **miracul-f donn-f apparir-ID4** à **I** vêtu-f de **couleur-f** très-**blanc-f**, au milieu de deux **gentil-f donn-f**, **a. r. être-ID3** de **+ long-f âg-m**; &, **passer-ID8** par un-f **voi-f**, **tourner-ID4 a.** yeux vers **I** & **I saluer-ID4** très **virtueus-m**, tant **r. I paraîtr-ID4** alors **voir** tous **a. term-m** de **a. béatitud-f**.¶

**Figure 6**
Encoding of original Italian text by Claude

The encoding of the original Italian text by Grok is definitely more precise and detailed than those of the other two, analyzing the text word by word.

**Table of Key Morphemes and Translations**

| Italian Morpheme | Pucci's Ideogram | Encoded Form (Partial) | Translation | Natural French | Explanation |
|---|---|---|---|---|---|
| **Ai miei occhi** | a M.-m occhi | a M.-m occhi | "À mes yeux" | "À mes yeux" | M.-m for possessive, correct preposition. |
| **Apparve** | apparire [remote past] | apparire [remote past] | "Apparut" | "Apparut" | Chart B for past tense. |
| **La gloriosa donna** | a. glorioso-f donna | a. glorioso-f donna | "La glorieuse dame" | "La glorieuse dame" | a. for article, f for feminine. |
| **Io la vidi** | I IIIf vedere [remote past] | I IIIf vedere [remote past] | "Je la vis" | "Je la vis" | I, IIIf for pronouns, past tense. |
| **Nobilissimo colore** | nobile-m couleur-m | di nobile-m couleur-m | "De noble couleur" | "D'une couleur très noble" | Superlative simplified, gender correct. |
| **Spirito della vita** | a. esprit-m di a. vita | a. esprit-m di a. vita | "L'esprit de la vie" | "L'esprit de la vie" | Literal, misses symbolic depth. |
| **Ella non parea** | IIIf non apparire [imperfect] | IIIf non apparire [imperfect] | "Elle ne semblait pas" | "Elle ne semblait pas" | Correct pronoun, imperfect tense. |
| **Colore bianchissimo** | couleur-m blanc | di couleur-m blanc | "De couleur blanc" | "D'une couleur très blanche" | Superlative simplified. |

**Updated Flowchart for Pucci's Translation Process**

**Python Code for Flowchart:**

python

```python
from graphviz import Digraph

# Initialize the flowchart
dot = Digraph(comment="Pucci's Translation Process for La Vita Nuova", format="png")

# Define nodes
dot.node("A", "Input: Ai miei occhi apparve la gloriosa donna...")
dot.node("B", "Parse morphemes: nouns, verbs, prepositions, etc.")
dot.node("C", "Apply Chart A: a., I, IIIf, M., r., &")
dot.node("D", "Apply Chart B: apparire [remote past], vedere [imperfect], etc.")
dot.node("E", "Handle adjectives/adverbs: glorioso-f, nobile-m")
dot.node("F", "Retain prepositions (di, a, da), nouns (donna, Beatrice)")
dot.node("G", "Dictionary lookup: yeux, dame, esprit, Béatrice")
dot.node("H", "Output: À mes yeux apparut la glorieuse dame...")

# Define edges
dot.edges(["AB", "BC", "CD", "DE", "EF", "FG", "GH"])

# Render the flowchart
dot.render("pucci_vita_nuova_flowchart", cleanup=True)

# Saves as 'pucci_vita_nuova_flowchart.png'
```

**Figure 7**
Encoding of original Italian text by Grok

Comparison of the excerpt from Dante translated by the three AIs:

| ChatGPT | Claude | Grok |
|---|---|---|
| À mes yeux apparut la glorieuse dame de mon esprit, celle que beaucoup appelaient Béatrice. Je la vis pour la première fois vers la fin de ma neuvième année. Elle apparut vêtue d'une couleur des plus nobles, ceinte et ornée de la manière qui convenait à sa très jeune âge. En ce moment-là, je dis en vérité que l'esprit de la vie commença à trembler si fortement qu'il semblait frémir jusque dans les moindres pulsations. Et je la voyais dotée de manières si nobles et louables, qu'on pouvait dire, comme l'a fait le poète Homère : "Elle ne semblait pas être la fille d'un homme mortel, mais d'un Dieu." Après que tant de jours furent passés, lors du dernier d'entre eux, il advint que cette merveilleuse dame m'apparut vêtue de la couleur la plus blanche, au milieu de deux dames nobles, plus âgées qu'elle ; et, passant par une rue, elle tourna les yeux vers moi et me salua très noblement, au point qu'il me sembla alors contempler les confins mêmes de la béatitude. | *À mes yeux apparut la glorieuse dame de mon esprit, laquelle fut par beaucoup appelée Béatrice. Je la vis presque dès la fin de ma neuvième année. Elle apparut vêtue de très noble couleur, ceinte et ornée à la façon qui convenait à son très jeune âge. En ce point je dis vraiment que l'esprit de la vie commença à trembler si fortement qu'il apparaissait dans les moindres pouls horriblement. Et je la voyais de si nobles et louables comportements, qu'on pouvait dire cette parole du poète Homère: Elle ne paraissait fille d'homme mortel, mais de Dieu.*<br>*Après que furent passés tant de jours, au dernier de ceux-ci, il advint que cette admirable dame apparut à moi vêtue de couleur très blanche, au milieu de deux gentilles dames, lesquelles étaient de plus long âge; et, passant par une voie, elle tourna les yeux vers moi et me salua très vertueusement, tant qu'il me parut alors voir tous les termes de la béatitude.* | À mes yeux apparut la glorieuse dame de mon esprit qui fut par beaucoup appelée Béatrice. Je la vis presque depuis la fin de mon année neuvième. Apparut vêtue de noble couleur, ornée à la manière qui convenait à son jeune âge. En ce point je dis véritablement que l'esprit de la vie commença à trembler si fort qu'il apparaissait dans les pouls minimes orriblement. Et je la voyais de si noble et louable portements que pouvait dire la parole du poète Homère : Elle ne semblait pas fille d'homme mortel mais de Dieu.<br>Puisque furent passés tant de jours, dans le dernier de ceci, arriva que la mirabile dame apparut à moi vêtue de couleur blanc, en milieu de gentilles dames qui étaient de plus long âge ; et, passant par une rue, tourna les yeux vers moi et me salua très vertueusement, tant que sembla à moi voir tous les termes de la béatitude. |

**Figure 8**
Comparison of the excerpt from Dante's *La Vita Nuova* respectively translated by ChatGPT, Claude and Grok

I began this experiment by selecting three AI systems freely available online to carry out the tests, including ChatGPT-4o and Grok 3. However, on August 7, 2025, OpenAI released its latest AI model, ChatGPT-5, and on August 11, 2025, xAI made available for free to all users the "Expert" version of Grok 4, which corresponds to a mode that systematically uses the Grok 4 model, even if this free access is time-limited and may include restrictions on usage volume for non-paying accounts. Hence the need to redo the tests using GPT-5 and Grok 4, following the same methodology, and to compare the new results with each other and with the previous ones.

### 3.1.1 ChatGPT 5 and Grok 4

**ChatGPT 5** gives this short explanation of the format:

- Each lexical item appears as either an ideogram (Pucci key) or a neutral dictionary stem (UPPERCASE). Example: GLORIOS-f = adjective "glorioso" marked feminine; OCCHI+ = eyes (plural).
- Verbs are shown as infinitive stems followed by the tense ideogram (e.g. APPARIRE +ID4 = apparve / passato remoto). When a verb is used as participle/adjectival form I show +ID9 (past participle) or +ID8 (gerund/present-participle).
- Passive voice in Pucci would be indicated by a passive marker plus agent: +PASS(DA-MOLTI) = "was called by many".
- The stream was kept compact because Pucci's machine would output a short, schematized representation that a target-language reader (with the dictionaries) would reconstruct and reorder to produce fluent output.

[a] [M.] [occhi+PL] [apparire{PST}] [a.] [glorios-f] [donna] [di] [M.] [mente] [r.] [chiamare{PST}+PASS(da molti)] [Beatrice] [l] [la] [vedere{PST}] [quasi] [a] [a.] [fine] [di] [M.] [anno] [nono] [apparire{PST}] [vestire-PART] [di] [colore] [nobile] [cinta] [&] [ornato-f] [in modo] [adatto] [a] [sua] [giovanissim-f] [età] [a.] [momento] [lo] [spirito] [di] [a.] [vita] [cominciare{PST}] [a] [tremare{INF}] [molto] [forte] [tanto_che] [apparire{IMP}] [nei] [menom+PL] [polsi] [orribilmente] [&] [l] [la] [vedere{IMP}] [con] [manier+PL] [si] [nobil-m/f] [&] [laudabil] [che] [potere{IMP}] [dire{INF}] [quella] [parola] [di] [a.] [poeta] [Omero] [:] [IIIIf] [non] [parere{IMP}] [figlia] [di] [uomo] [mortale] [ma] [di] [Dio] [poi_che] [passare{PST}] [tanto] [giorno+PL] [in] [a.] [ultimo] [di] [questo+PL] [avvenire{PST}] [che] [questa] [mirabil-f] [donna] [apparire{PST}] [a] [l] [vestire-PART] [di] [colore] [bianchissim-f] [in mezzo di] [due] [gentil-f] [donna+PL] [r.] [essere{IMP}] [di] [più] [lung-f] [età] [&] [passare{GER}] [per] [una] [via] [volgere{PST}] [occhio+PL] [verso] [l] [&] [salutare{PST}] [l] [molto] [virtuosamente] [tanto_che] [parere{PST}] [vedere{INF}] [tutto+PL] [a.] [termine+PL] [di] [beatitudine]

Pucci Raw French Output

À mes yeux apparut la glorieuse femme de mon esprit, laquelle fut appelée par beaucoup Béatrice ; je la vis presque à la fin de mon neuvième année ; apparut vêtue de couleur noble, ceinte et ornée à la façon que à sa très jeune âge convenait ; en ce moment je dis que l'esprit de la vie commença à trembler si fortement que apparaissait dans les plus petits poignets horriblement ; et je la vis avec comportements si nobles et louables que se pouvait dire cette parole du poète Homère : elle ne paraissait pas fille d'homme mortel mais de Dieu ; après que passèrent tant de jours, dans le dernier de ceux-ci, arriva que cette admirable femme apparut à moi vêtue de couleur très blanche, au milieu de deux femmes gracieuses qui étaient de plus longue âge ; et passant par une rue, tourna les yeux vers moi et me salua très vertueusement ; si bien qu'il me parut alors voir tous les termes de la béatitude.

**Encoding and translation by ChatGPT 5**

**Figure 9**
Pucci-style ideogram stream (machine output) and Pucci raw French output by ChatGPT 5

The translation process by **Grok 4** simulates the "machine" output directly in simplified French (basic French words with approximate structure), as if the recipient had partially rearranged it. But to explain in detail, it breaks it down phrase by phrase:

- Original Italian
- Input Simplification (direct and reduced Italian)
- Encoding with Ideograms (neutral interlanguage, using basic Italian/French common words + symbols)
- Decoding/Reconstruction in French (lexical equivalents via simulated dictionaries, rearrangement for "good enough")
- Final Result (as in my previous response, with explanations of choices)

The result begins with "Aux mes yeux", which is very strange since until now all previous translations began with "À mes yeux", justifying this choice in the following way:

- Articles/Pronouns: "A" (preposition + article) → a. (article) + M. (possessive my).
- "Aux mes yeux": simplification of "a. M. occhi+" → "Aux" (preposition + plural article) + "mes" (M. for my + plural).

A rudimentary error that Pucci himself explicitly dismissed. As he observed, even if the user is unable to identify the precise cause of the problem, it is immediately evident that "aux mes" is not acceptable French and that the fault lies in the form "aux". By consulting the *differential tables*, they show that Italian "a" drops before a possessive and articulated forms reduce to simple ones; hence "aux" is corrected to "à." (However, these tables only apply to Romance languages).

Below is the translation according to the principles of Federico Pucci's method (based on the provided ideograms and universal keys):

> Aux mes yeux apparut la glorieuse femme de ma mente, la quelle fut par beaucoup appelée Beatrice. Je la vis presque de la fin de mon année neuvième. Elle apparut vêtue de couleur très noble, ceinte et ornée à la manière qui à sa âge très jeune se convenait. En ce point je dis véritablement que l'esprit de la vie commença à trembler si fortement qu'il apparait dans les moindres pulses de façon horrible. Et je la voyais de si nobles et louables comportements, que l'on pouvait dire cette parole du poète Homère : Elle ne paraissait pas fille d'homme mortel, mais de Dieu.
>
> Puis que furent passés tant de jours, dans le dernier de ces, arriva que cette merveilleuse femme apparut à moi vêtue de couleur très blanche, au milieu de deux gentilles femmes, les quelles étaient de âge plus longue ; et, passant par une voie, tourna les yeux vers moi et me salua très vertueusement, tant que il me parut alors voir tous les termes de la béatitude.

**Figure 10**
Pucci-style French output by Grok 4

Overall, the number of removals and additions remains in line with the average of previous results, which this heat map perfectly sums up.

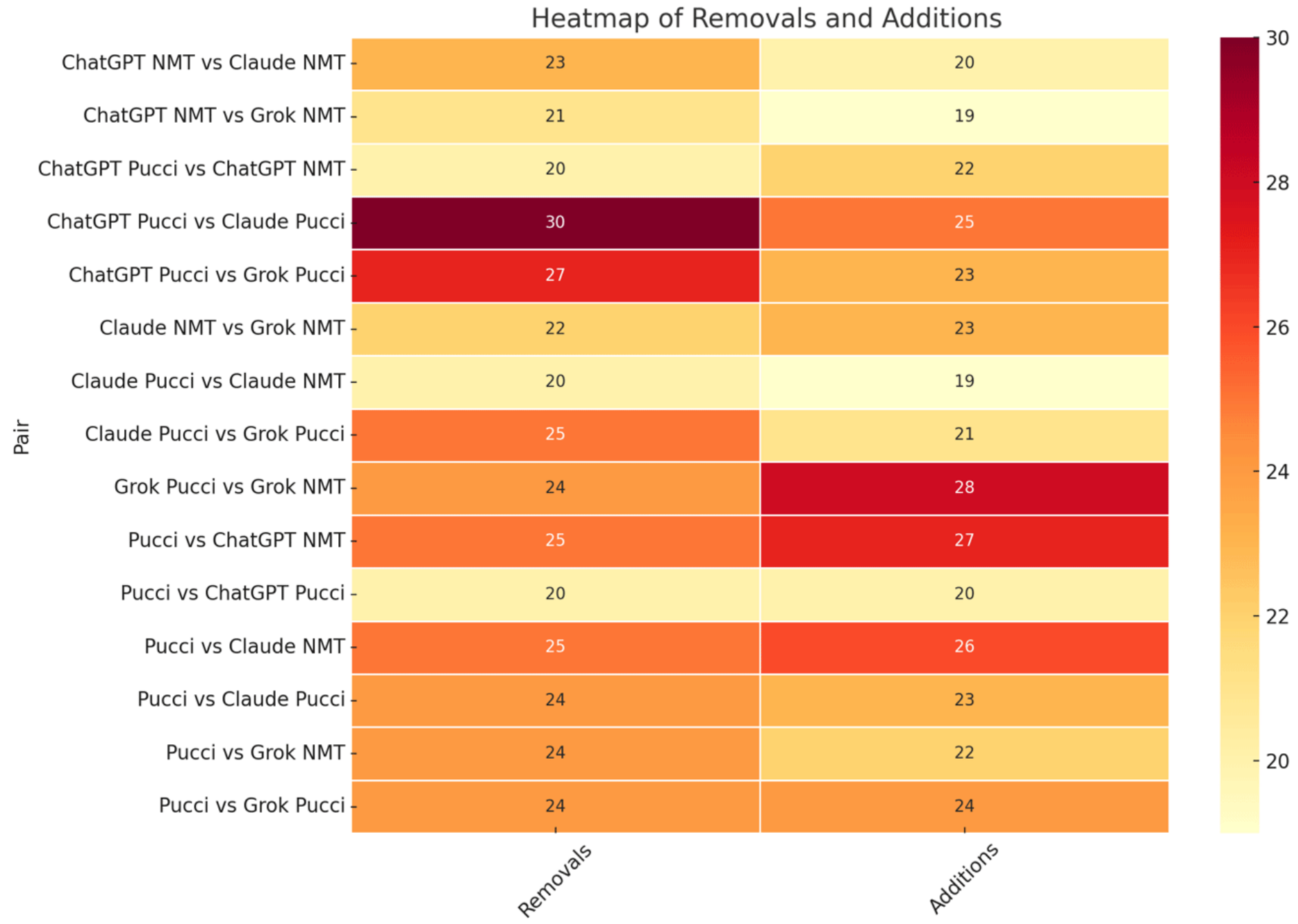

**Figure 11**
Heatmap of removals and additions between versions A and B

Within the framework of textual comparison between versions A and B, this study defines removals as linguistic components (one word or a few words) present in the source version A yet absent or transformed in version B, and additions as components manifest in version B but lacking in version A, acknowledging that such variations may encompass elements as minor as punctuation marks.

### 3.2 Control test between results produced by the first AI models tested and between those and Pucci's output

We had initially fifteen possible comparisons, divided into three groups:

Group 1 – Current AI-based NMT systems (1–3)

1. ChatGPT NMT vs. Claude NMT – 23 removals + 20 additions
2. ChatGPT NMT vs. Grok NMT – 21 removals + 19 additions
3. Claude NMT vs. Grok NMT – 22 removals + 23 additions

Group 2 – AI Models trained with Pucci's method vs. AI NMT (4–9)

4. ChatGPT Pucci vs. Claude Pucci – 30 removals + 25 additions
5. ChatGPT Pucci vs. Grok Pucci – 27 removals + 23 additions
6. Claude Pucci vs. Grok Pucci – 25 removals + 21 additions
7. ChatGPT Pucci vs. ChatGPT NMT – 20 removals + 22 additions
8. Claude Pucci vs. Claude NMT – 20 removals + 19 additions
9. Grok Pucci vs. Grok NMT – 24 removals + 28 additions

Group 3 – Pucci's original outputs vs. AI systems (both Pucci-trained and NMT) (10–15)

10. Pucci vs. ChatGPT Pucci – 20 removals + 20 additions
11. Pucci vs. Claude Pucci – 24 removals + 23 additions
12. Pucci vs. Grok Pucci – 24 removals + 24 additions
13. Pucci vs. ChatGPT NMT – 25 removals + 27 additions
14. Pucci vs. Claude NMT – 25 removals + 26 additions
15. Pucci vs. Grok NMT – 24 removals + 22 additions

Detailed results are presented in Appendix G (Figure G.1).

☞ Since the comparisons with ChatGPT 5 and Grok 4 versions are in line with these results and do not add anything more significant to the discussion, I felt it was unnecessary to detail them in the context of this article.

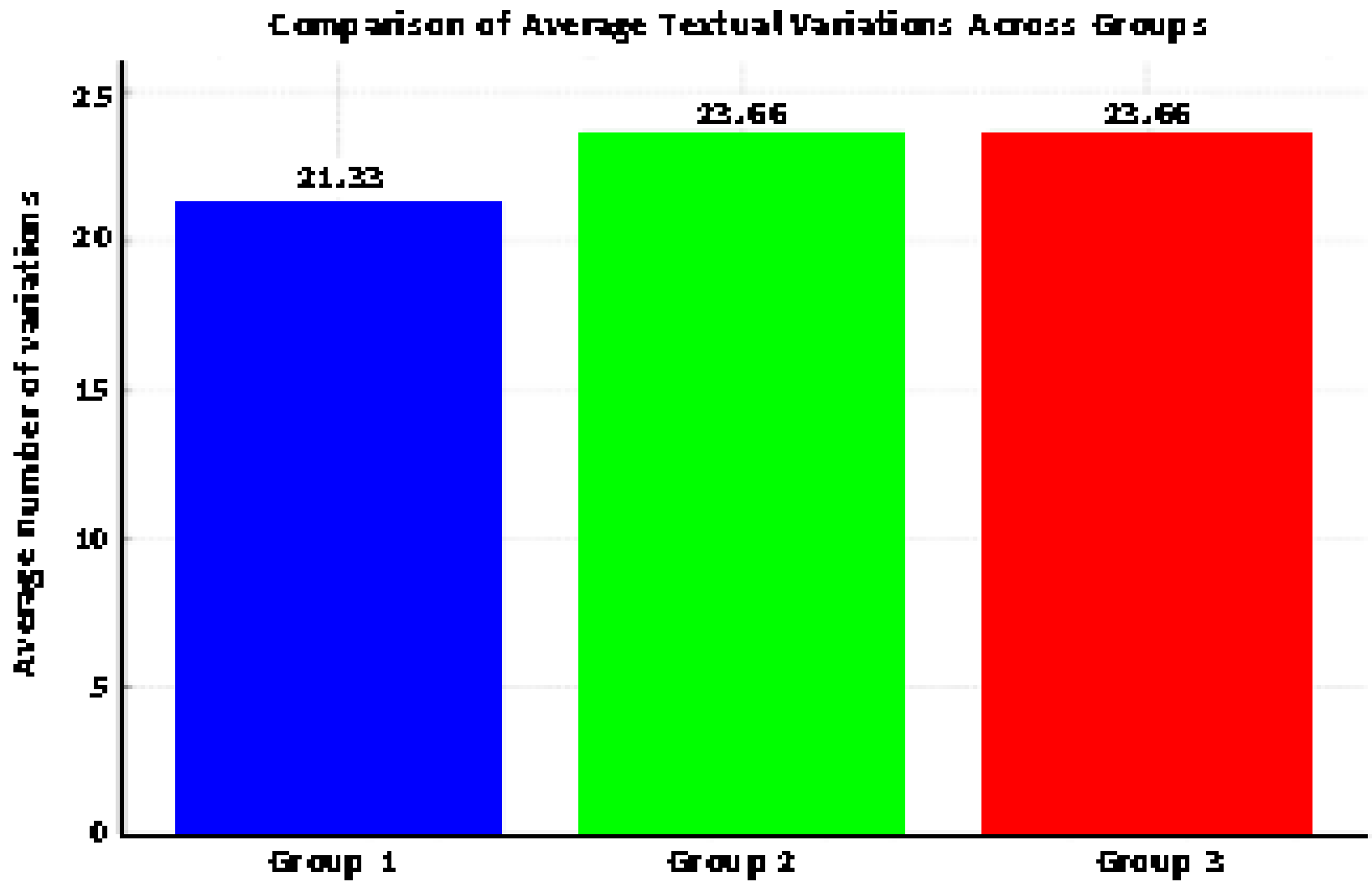

**Figure 12**
Comparison of Average Textual Variations Across Groups 1 – 2 – 3

**Quantitative comparison and interpretation of translation outputs**

**1. Low to moderate divergence within the same method across AIs**
When comparing outputs from different AI systems applying the same strategy (e.g., Claude NMT vs. Grok NMT), divergence is modest. For instance, Claude NMT vs. Grok NMT showed 22 removals and 23 additions, while Claude Pucci vs. Grok Pucci showed 25 removals and 21 additions. These results indicate that, although phrasing differs, the shared method exerts a unifying effect across systems.

**2. Low to moderate divergence within the same method across AIs**
Comparisons between Pucci's symbolic method and standard NMT within the same AI yield slightly higher divergence. For example, ChatGPT Pucci vs. ChatGPT NMT showed 20 removals and 22 additions, while Claude Pucci vs. Claude NMT showed 20 removals and 19 additions. This suggests that methodology strongly shapes the translation output, though each AI retains a consistent stylistic "signature."

**3. Highest divergence between Pucci's original and modern AI outputs**
The greatest differences occur when Pucci's 1931 translations are compared with modern NMT outputs: Pucci vs. ChatGPT NMT (25 + 27), Pucci vs. Claude NMT (25 + 26), and Pucci vs. Grok NMT (24 + 22). Each yields roughly 50 changes, underscoring that Pucci's symbolic framework produces a distinct stylistic and structural pattern not replicated by neural methods.

**Summary by Comparison Type**

- NMT vs. NMT across AIs: ~45 changes. Despite a shared foundation, models diverge in wording, reflecting subtle architectural differences.
- Same AI, different methods (NMT vs. Pucci-trained): ~42 changes. Methodology introduces noticeable structural variation, even within the same model.
- Same method, different AIs (e.g., Pucci across Claude vs. Grok): ~45–50 changes. Divergence reflects differences in each model's internal logic.
- Pucci original vs. any AI translation: ~48–52 changes. Pucci's symbolic system remains distinct from both neural and AI-simulated Pucci outputs.

**Interpretation**

* Group 1 (AI NMT only, comparisons 1–3): lowest variation, average 21.33, showing strong internal consistency across contemporary neural systems.
* Groups 2–3 (Pucci-trained AIs and Pucci original, comparisons 4–15): higher variation, identical average 23.66.
This systematic elevation (+2.33, ≈11%) shows that introducing Pucci's framework, whether directly or via AI, consistently increases divergence from standard NMT.
The effect is medium in size (Cohen's $f = 0.29$, $\eta^2 = 0.077$), meaning group membership explains about 8% of the variance in editing operations. While the difference is not dominant, it is stable and measurable, pointing to the methodological distinctiveness of Pucci's system.

In conclusion, Pucci's framework, as detailed in Section 2.2, produces coherent outputs distinct from standard NMT, with AI replication (Groups 2 and 3) preserving its defining characteristics (see Appendix G, Figure G.1 for metrics). This underscores its historical significance, further explored in Section 3.3.3.

### 3.3 BLEU scores and other metrics (chrF, METEOR)

Since BLEU (*Bilingual Evaluation Understudy*) [12] is designed for same-language comparisons, based on a "gold standard" text, the first step was to understand whether Pucci's translation was of sufficient quality to serve as a reference, given that the metric is only as reliable as the text against which you compare a candidate translation.

A gold standard is a translation considered sufficiently accurate, fluent, and representative of the source text to serve as the benchmark for automatic evaluation. BLEU does not “know” about meaning or style: it just compares the n-grams in the candidate translation to those in the reference.

So, to ensure a robust evaluation of Pucci's translation, we adopted **a two-part methodology:**

**Firstly**, we performed a qualitative analysis using a **literary translation evaluation grid** to assess the translation's quality across several key criteria, including its fidelity to the original's meaning, its clarity and fluency in French, its style and musicality, its adherence to the source text's register and tone, and the creativity and precision of its lexical choices; then, we compared these results with those of three royalty-free translations of the same passage from *La Vita Nuova*, respectively translated by Maxime Durand-Fardel (1898) [1], Hippolyte Godefroy (1901) [2] and Henry Cochin (1905) [3].

This approach is consistent with the principles of open access, given that the more modern French versions produced by renowned translators such as Jacqueline Risset, André Pézard or Christian Bec remain under copyright and are therefore not freely available. Nevertheless, a comparison between these authoritative translations and Pucci’s version would be of considerable scholarly interest.

So, each translation presents its own specificities, with distinctive strengths and limitations; Pucci’s "good enough" version, while not without imperfections, legitimately asserts its place within the broader corpus of these French renderings of Dante’s *Vita Nuova*, and may thus be considered a suitable gold standard for further analysis, particularly through the attribution of a BLEU score to the various texts.

**Secondly**, we proceeded to calculate and compare the **BLEU scores** of Pucci's version and these three candidate literary translations. As we have already seen, the translated French text by Pucci can be found on pages 10 and 11 (see Appendix H, Figure H.1 for Italian excerpt).

#### 3.3.1 The literary translation evaluation grid

Scoring out of 10 for each criterion:

- Fidelity to Meaning
- Clarity and Fluency in French
- Style and Musicality
- Adherence to Source Text, Register and Tone
- Creativity / Lexical Choices

we can then giving an average final score (which is the mean of the five criteria scores for each translation).

Below are a comparison table and a global analysis synthesizing the four evaluations into a broader perspective.

| Translator | Fidelity to Meaning (/10) | Clarity and Fluency in French (/10) | Style and Musicality (/10) | Adherence to Source's Text, Register and Tone (/10) | Creativity / Lexical Choices (/10) | Average Final Score (/10) |
|---|---|---|---|---|---|---|
| Durand-Fardel (1898) | **7** (Takes more liberties, but captures essence) | **8** (Much smoother and idiomatic French, easier to read) | **7** (More literary rhythm, softer cadences, and some elegance) | **8** (Maintains elevated, solemn register with some grace ("courtoisie infinie") | **8** (Shows interpretive choices that enhance readability while remaining close to Dante's mood) | **7.6** |
| Godefroy (1901) | **7** (Faithful overall, but with romanticized additions and omissions) | **8** (Smooth, idiomatic, occasionally convoluted) | **7** (Elegant yet over-embellished; cadence leans rhetorical rather than lyrical) | **7** (Sentimental, modern-romantic inflection, less restrained than Dante) | **8** (Inventive lexical choices, though distant from original texture, with some omissions) | **7.4** |
| Cochin (1905) | **9** (Very close to Dante, minimal interpretive liberties) | **7** (Archaic phrasing, occasionally cumbersome, but still intelligible) | **8** (Retains more of the medieval solemnity and cadence) | **9** (Successfully preserves Dante's elevated, almost sacred tone) | **7** (Faithful and sometimes elegant, though occasionally stiff and dated) | **8.0** |
| Pucci (1931) | **9** (High fidelity; minor omissions or some weak rendering ("très petits pouls horriblement") | **7** (Fluent but mechanical and dated phrasing, e.g., "comme il se convenait") | **6** (Lacks rhythm, overly prosaic, flat cadences) | **7** (Maintains reverent tone but without the poetic solemnity of Dante) | **6** (Practical choices like "pensée" for "mente"; functional but not highly inventive) | **7.0** |

**Figure 13**
Comparison of criteria and scoring of each translator

When compared collectively, the four translations reveal distinct priorities in the art of rendering Dante's *Vita Nuova* into French. Durand-Fardel's version privileges clarity and fluency, embracing a freer and more interpretive method that enhances elegance, readability, and literary appeal, even if this occasionally introduces nuances absent from Dante's text. Godefroy, by contrast, moves further toward embellishment, adopting a romanticized and lyrical register that softens the medieval austerity of the original. While this risks diluting Dante's theological and symbolic density, it also demonstrates the creative possibilities of interpretive freedom, opening the text to contemporary sensibilities. Cochin, situated between these approaches, achieves perhaps the most accomplished balance: rigorously faithful to Dante's meaning while preserving the elevated register and solemn cadence that convey the spiritual intensity of the work. Finally, Pucci stands out for its literalism, providing a close textual witness of Dante's words but at the cost of stylistic resonance and poetic vitality.

Indeed, the comparative analysis suggests that no single translation fully encapsulates Dante's genius, but each sheds light on different facets of the text. Durand-Fardel conveys accessibility and elegance; Godefroy reveals the potential of imaginative re-creation; Cochin emphasizes fidelity and solemnity; and Pucci offers a precise, if overly rigid, textual record. Taken together, these translations underscore the inherent tension in literary translation between literal accuracy and aesthetic rendering; they illuminate the multiplicity of interpretive choices that shape the reception of Dante in the French literary tradition.

Moreover, if we consider that Pucci's goal was not literary excellence but accessibility for readers not knowing foreign language, the fact that from a philological standpoint his translation may appear overly literal, flat in rhythm, and stylistically awkward, is not a failure, but rather a different kind of success. By privileging transparency of meaning over poetic eloquence, his work, which anticipated almost a century ago the modern distinction between literary translation and functional translation—the latter emphasizing accessibility, user-friendliness and comprehension rather than aesthetic fidelity—is today entirely suitable as a gold standard.

### 3.3.2 Comparative analysis using BLEU score, chrF metric and METEOR test

Although the score of 8.0 would conventionally serve as the reference translation, in this study we adopt an inverse approach to evaluate the influence of Pucci's pre-digital method. While BLEU is a widely used automatic evaluation metric in the field of machine translation (MT), it is rarely applied to human translation (HT). Nonetheless, employing BLEU to compare machine-generated outputs with human-produced translations can yield valuable insights, particularly for analytical and research-oriented purposes.
Using Grok 4, we calculated the BLEU scores of Pucci's translation and compared them against the three previously discussed literary translations and additionally included as a candidate the neural machine translation (NMT) output generated by ChatGPT-5. Originally proposed by Papineni et al. [12], the BLEU metric measures the degree of similarity between a candidate translation and a reference text by computing the overlap of n-grams—defined as contiguous sequences of one to four words—and applying a length-based adjustment known as the brevity penalty. Higher BLEU scores indicate greater lexical correspondence with the reference. However, because the metric privileges exact surface matches, it may undervalue stylistic variation or creative reformulation, a limitation that is particularly relevant in the evaluation of literary translations (see Appendix I for all candidate's translations).
Due to the limitations of BLEU, translation studies increasingly adopt a multidimensional approach, supplementing BLEU with metrics such as chrF (Character-based F-score) [18] and METEOR (Metric for Evaluation with Explicit ORdering) [17]. The chrF metric offers a complementary perspective by operating at the character level, where it captures morphological and grammatical accuracy. METEOR introduces a further dimension by incorporating synonym recognition, stemming, and penalties for word-order disruption. Finally, the inclusion of ChatGPT-5 as a control case underscores the complementarity of the evaluation metrics and situates Pucci's work within a broader historical perspective. The system's very high scores—67.45 BLEU, 85.60 chrF, and 82.62 METEOR—indicate a strong level of both formal and semantic alignment. These results suggest that modern AI translation approximates Pucci's literalist method with striking fidelity, whereas human translators such as Durand-Fardel and Godefroy diverge significantly due to interpretive expansions and stylistic interventions. This observation not only enriches the analysis but also sharpens our understanding of Pucci's position in the history of translation. Here is the updated table:

| Translation Method | Corpus BLEU | chrF | METEOR | 1-gram Precision (chrF) | 6-gram Precision (chrF) | Precision (METEOR) |
|---|---|---|---|---|---|---|
| Godefroy (1901) (~7.4) vs. Pucci (1931) | 12.47 | 63.81 | 56.76 | 94.59% | 64.86% | 62.16% |
| Fardel (1898) (~7.6) vs. Pucci (1931) | 17.34 | 66.87 | 59.85 | 95.45% | 67.42% | 65.45% |
| Cochin (1905) (~8.0) vs. Pucci (1931) | 46.23 | 75.31 | 78.43 | 97.09% | 74.20% | 83.64% |
| ChatGPT 5 (2025) vs. Pucci (1931) | 67.45 | 85.60 | 82.62 | 98.29% | 81.77% | 88.10% |

**Figure 14**
Comparison of BLEU, chrF and METEOR scores

### 3.3.3 Interpretation

First, the exceptionally high alignment between Pucci and the unconstrained ChatGPT-5—achieved without any awareness of his rules—demonstrates that large-scale neural translation systems, trained on vast corpora to detect statistical patterns, have independently developed an approach that converges conceptually with Pucci's symbolic and interlingual model. The mechanisms differ radically—statistical learning versus mechanical formalism—but the results reveal a profound affinity between Pucci's theoretical intuitions and contemporary machine translation.

Second, the comparison with human translators highlights this divergence further. The relatively low scores of Durand-Fardel and Godefroy, especially when contrasted with both Pucci and ChatGPT-5, confirm that their methods were firmly rooted in the humanist tradition of interpretive and stylistic mediation. Such approaches, while literary in their own right, are fundamentally distinct from the paradigm of mechanical or machine-like translation pursued by Pucci and approximated, albeit differently, by AI systems.

Cochin, by contrast, occupies an intermediate position. His BLEU score of 46.23, combined with high chrF (75.31) and METEOR (78.43), indicates a method closer to Pucci's literalism. Cochin preserved morphological detail and semantic clarity more consistently than his contemporaries, while still allowing for stylistic variation. His work shows that literalist tendencies were not absent from early twentieth-century human translation, but his method lacked the systematic rigor that defines Pucci's quasi-mechanical approach. In this sense, Cochin functions as a bridge: closer to Pucci than Fardel and Godefroy, but still firmly within the humanist paradigm of interpretive translation.

The decisive evidence emerges, however, in the inclusion of ChatGPT-5 as a control case: this unconstrained AI model produces results that most closely approximate Pucci's *verbum pro verbo* approach. This value, by far the highest in the entire dataset, serves as the centerpiece of the analysis. It demonstrates conceptual convergence: Pucci's early twentieth-century "interlingual" system of symbolic equivalences parallels the statistical and neural methods of modern AI translation, even though the underlying mechanisms differ. Without being trained on Pucci's method, ChatGPT-5 nevertheless generates output that echoes his style, structure, and vocabulary.

In this light, the comparative results offer a threefold conclusion. First, human translators such as Fardel and Godefroy diverged from Pucci through interpretive freedom, while Cochin preserved a partial alignment with his literalist ethos. Second, current AI approximates Pucci more closely than any human translator, confirming the kinship between his method and machine translation. Third, as established in Section 2.2 and evidenced by the metrics in Section 3.2, Pucci's symbolic system positions him as a pioneering precursor to modern machine translation, his interlingual approach anticipating computational methods by nearly a century.

In conclusion, the complementarity of BLEU, chrF, and METEOR not only enriches the technical analysis but also situates Pucci in the broader history of translation. His work appears less as an eccentric anomaly and more as a visionary anticipation of computational approaches—so structurally rigorous that, although today's data-driven models are not rule-based, they can still be steered to reproduce aspects of his system when guided by carefully designed inputs.

## 4. Discussion

The remarkable convergence between original and reproduced outputs—one generated by a human translator in 1931 using a symbolic, manual system, and the other by artificial intelligence 94 years later following that identical methodological framework—represents a significant achievement in the field of historical reconstruction, where the loss of technical detail frequently impedes high-fidelity replication, particularly within language technologies.
While the words differ slightly in formulation or syntax, the semantic core and sentence structure remain remarkably similar across both translations. The observed variations primarily relate to stylistic choices or synonym substitutions, such as "nobilissimo" rendered as "très noble" versus "d'une noblesse extrême," alongside minor differences in pronoun placement reflecting contemporary French conventions. Notably, gender markings and adjective agreement are consistently applied in both versions, demonstrating systematic adherence to grammatical principles.
The overall structural alignment between the two versions provides compelling empirical confirmation that the AI was not merely imitating general translation patterns, but rather faithfully reproducing Pucci's underlying logical framework. This adherence is evidenced through the AI's strict application of Pucci's systematic approach: interpreting each sentence step-by-step using symbolic keys and ideograms, generating Italian output through symbolic rules rather than relying on web-based translations, and implementing the complete methodological sequence of morphological reduction of the input language, transformation into ideograms and grammatical keys, and recomposition in the target language following Pucci's syntax-flexible principle.
Minor variations between the translations include subtle differences in verb tense selection and word order—variations typical in French-Italian translation practices—yet both versions preserve proper names, rhetorical flourishes, and sentence rhythm with exceptional fidelity. Perhaps most significantly, this suggests a level of abstraction and formalism in Pucci's original system that rivals modern symbolic approaches in AI — and, most notably, that his work anticipated core elements of machine translation decades ahead of the Translation memorandum by Warren Weaver and the IBM–Georgetown experiment.

In a 1967 article, *The Georgetown-IBM Experiment of 1954: An Evaluation in Retrospect*, Paul Garvin [7] (a linguist on the Georgetown team) pointed out:

- the disproportionate enthusiasm generated by this experience,
- the non-representative nature of the demonstration,
- its limited transferability to other domains or languages,
- and its lack of real linguistic depth.

Considering these critiques, it becomes plausible — even historically credible — to imagine a different trajectory for MT, one based on the rigorous, elaborated system Pucci had already published 25 years earlier.

Rather than building an artificial demo with a few dozen phrases, IBM and Georgetown might have developed a genuinely generalizable interlinguistic prototype, opening the door to an early, symbolic and explainable form of machine translation — modular, adaptable, and expandable through domain-specific dictionaries.

Had Pucci's work been known at the time, it might have significantly influenced the direction of MT research. This is not about rewriting the past but about completing it—doing justice to an invention that was never exploited but has now been empirically validated. If Pucci's method had been used or known before the 1950s, it's likely the technological stagnation denounced in the 1960s (especially in the ALPAC report) [13]—which led to widespread disillusionment—might have been avoided or at least softened. Instead of abandoning symbolic approaches in favor of purely statistical or neural models, the MT field could have progressed with a hybrid foundation, incorporating rules and interlingua logic from the outset. Ironically, many of these principles were "rediscovered" decades later in the 1990s and 2000s.

Modern AI has validated Pucci's system, with extremely close results to those he produced manually in 1931. With a low divergence between human and AI outputs, this experiment demonstrates the stability, reproducibility, and underlying formal coherence of Pucci's method. This positions his work not as outdated but technologically relevant — perhaps even more so today, in an age seeking transparent, explainable, and symbolic alternatives to opaque neural networks.

This study repositions Federico Pucci not as a failed or marginal figure, but as an emblematic figure of the "too-far-ahead-of-their-time" inventor, whose contributions were erased not because of technical inferiority, but because of structural neglect. Factors such as geographic isolation, lack of institutional support, and a non-commercial agenda led to his systematic exclusion from the historical canon of machine translation.

Technological history is not inherently meritocratic, but shaped by social, economic, and institutional forces. Pucci is presented as a case study in how the "sociology of innovation" often suppresses those who innovate outside dominant centers of power.

The rediscovery of Pucci's method is framed as not only a historical correction but a call to action. It asks:

- How many other innovations have been lost due to systemic blind spots?
- How can we recover and reevaluate them?
- Can symbolic systems, like Pucci's, offer more inclusive and explainable tools in contemporary AI?

The text also points to future research: on one side, in the final part of his work, Pucci introduces a derivative approach to translating French texts without any knowledge of the language and proposes a general theory that has not been pursued since, a research avenue that would merit closer examination today; on the other side, testing the adaptability of Pucci's symbolic method across non-Romance languages, to see if it offers a universalizable approach to translation. This isn't just about recovering the past — it's about reclaiming lost futures.

## 5. Rewriting the origins of Machine Translation

Thus, the story of Federico Pucci makes it possible to revise the origins of Machine Translation (MT) by offering a powerful counter-narrative to the dominant, institutionalized history of the field.

Pucci's method wasn't just early — it was radically different in its intent and design. He proposed a universal, ideographic interlingua that could serve as a bridge across Romance languages and potentially others. This predates the formal development of symbolic systems in MT and resembles the interlingua architectures used in later Rule-Based Machine Translation.

This also rewrites the timeline. MT didn't begin in the 1950s with computers and modern algorithms, nor with IBM, but quietly in Italy in 1929, with Federico Pucci, who skillfully organized mental structures into systems:

**1929** (December): Federico Pucci presents his study on the "mechanical translator" for the first time in Salerno.
**1930**: Federico Pucci's participation in the first National After-Work Arts and Crafts Exhibition of Bolzano – literary section, with his concept of "mechanical translator", awarded a silver medal.

**1931**: Federico Pucci publishes in Salerno what we might consider to be the first book ever published anywhere on a "mechanical translating device", called: "*Il traduttore meccanico ed il metodo per corrispondersi fra Europei conoscendo ciascuno solo la propria lingua: Parte I.*"
**1932**: likely construction of a prototype "translating machine" by Georges Artsrouni, later destroyed. No document has been kept about it, except for a photograph that makes a description impossible.
**1932**: Warren Weaver becomes director of the Rockefeller Foundation.
**1933**: filing of patent and presentation to Soviet authorities of Petr Petrovič Smirnov-Trojanskij's machine, probably at the design and description stage. (Source)
**1933-1935**: filing and grant of patent to Georges Artsrouni on 22 July 1933 and construction of his "mechanical brain"
**1935**: presentation of Federico Pucci's "mechanical translator" at the Inventors Competition, part of the Trade Fair of Paris, receiving a silver medal for a "a method for translating languages without knowing them"! (Source)
**1937**: Georges Artsrouni presents some machines at the National Exhibition of Paris, the principle of which received a Grand Prix award for mechanical data processing, according to the inventor himself.
**1939-1945**: World War Two

**1949**

- Around **May 10**, Federico Pucci sent his first registered letter to the U.S. presidency in which he submitted a booklet which translates Italian into every language, requesting funding for his invention, without knowing anything yet about any "electric brain";
- Around **May 31**, he read in an Italian newspaper (unspecified) the first United Press dispatch (traced in several newspapers, not just American, in early June 1949), announcing the construction at the University of California, Los Angeles, of an "electric brain" by the Bureau of Standards, capable of translating foreign languages;
- On **June 13**, Pucci sent his second registered letter to President Truman (received on July 25), containing a booklet which translates French and called the "electric translation";
- On **July 10**, Pucci sent his first (long) letter to the Italian CNR (National Research Council), in which he details very precisely the circumstances which led him to conceive his invention and in which he claims his anteriority over the American "electric brain"; the CNR answered on July 20;
- On **July 15**, Warren Weaver published his famous memorandum entitled "Translation";
- On **August 26**, a United Press dispatch announced Federico Pucci's invention, 20 years after its first public presentation (December 1929!);
- **During the year**, Pucci published 4 of the 10 books he dedicated to his invention.

**1950**: On October 17, Pucci sent his second letter to the Italian CNR (National Research Council); the CNR answered on October 27.
**1954**: January 7, the Georgetown–IBM experiment took place at the New York headquarters of IBM.
**1955**: William Nash Locke and Andrew Donald Booth published "Machine Translation of Languages", in which they reproduced the Weaver's memorandum.
**1960**: Federico Pucci published his tenth and latest book, entitled Il traduttore dinamo-meccanico: Serie A. L'invenzione per la traduzione immediata e rapida nelle lingue dell'Occidente senza conoscerle e quasi senza vocabolario... Tedesco - Italian [The dynamic-mechanical translator: Series A. The invention for the immediate and rapid translation in Western languages without knowing them and practically without a dictionary... German - Italian], 30 years after the first dedicated to his invention.
**1997**: John Hutchins find traces of the second United Press dispatch and mentioned Pucci's name in two different documents [7, 12] (see Appendix C, Figure C.1).
**2017**: I decided to create an infographic on the history of machine translation and I came across these two snippets from Hutchins. It's the origin of all of this story...

## 6. Conclusion

In a world saturated with linguistic technologies and real-time translation systems available freely online, it is paradoxical that one of the very first thinkers of such a system—Federico Pucci—remains a total unknown. The historical silence surrounding his foundational contribution raises two crucial questions: how could such oblivion occur and persist? What does it reveal about the way we write the history of innovation?

Despite the cognitive and technical limits of his time, Pucci nonetheless thought and designed a symbolic interlingua—crafted by hand—composed of ideograms and "universal" keys encoding the grammatical and semantic meaning of a sentence before re-expressing it in the target language. His method demonstrates remarkable theoretical depth and methodological coherence.

This study's central contribution is to empirically validate this prescient design by treating our research as a **proof of concept**. The project's foundational question was whether a past, never-realized mechanical translation system could be robust enough to be replicated and validated by modern artificial intelligence. As a demonstration, we utilized AI to re-translate the passage from Dante according to Pucci's detailed rules. The successful replication, evidenced by the remarkable convergence between the AI's translation and Pucci's original, constitutes the proof. It demonstrates that his method was not merely a theoretical curiosity, but a coherent, formalizable and reproducible system—a full century before the technology existed to implement it. This finding alone serves as a powerful asset to the historical record.

This highly structured system not only prefigures RBMT (Rule-Based Machine Translation) architectures but also anticipates some of today's computational interlingua models—abstract representations used in symbolic or hybrid MT systems to mediate between source and target languages. This is already a hybrid human / machine translation + post-editing (MTPE) workflow that is way ahead of its time, demonstrating the prescience of Pucci's design nearly a century before modern AI could bring it to life.

Pucci's story—far from being a mere antiquarian curiosity—challenges our collective responsibility as researchers, engineers, linguists, and historians. How can we better preserve, explore, and revalue forgotten knowledge? As AI today tends toward convergence and uniformity, might it not be time to reopen forgotten pathways, revisit symbolic systems, and reimagine how we approach language technologies?

Pucci's prematurely abandoned invention, deeply inclusive and universalist, breaks from the linguistic elitism still often present today. Thanks to modern AI, it finds unexpected new relevance, and **we can now experiment by translating like a century ago**. This is not a matter of nostalgia, but a source of innovation.

Despite this, Pucci remains entirely ignored by academia—a victim of the systemic exclusion of innovations that lack institutional backing. Yet his rediscovery in 2017, *triggered by a double mention in John Hutchins' archives*, prompts a critical review of his absence from MT history and opens the way for a new, broader narrative.

Bringing Pucci back into conversation is not merely an act of historical justice. It is an opportunity to rethink what we value in technical innovation—and to ensure that the future we build includes the many voices that history too often leaves behind.

## 7. Acknowledgements

The author warmly thanks Oriana de Majo, granddaughter of Federico Pucci, for his encouragement and for having greatly helped me to continue this work whenever I was at an impasse, most often due to the lack of documentation, and for her careful proofreading of my article.
The author gratefully acknowledges John Hutchins for being the first—and only—person to mention Federico Pucci in one of his many research projects. I tried twice (on April 11, 2018, and on March 12, 2019) to contact him by email (@hutchinsweb.me.uk) to let him know about the great human adventure behind the story of Federico Pucci, but I imagine he never read my emails.
This work stems from entirely independent research without any institutional affiliation, initiated on March 12, 2017. It was only on July 26, 2025 — more than eight years later — that I discovered the potential for operationalizing Pucci's method within contemporary AI systems to produce accurate translations, based on his invention dating back to the 1930s. This comprehension emerged when an AI system remarked: "*The lack of access to Pucci's original materials and actual charts, or to a complete 'book-machine' prototype, together with missing details on ideogram mappings, rule application, and chart structure, complicates implementation…*"
The author extends gratitude to the support of open-access AI systems and publicly available translation resources that made this investigation possible.
The author also wishes to thank the Truman Presidential Library for their assistance in confirming they indeed received a registered letter from "Frederic [sic] Pucci."
Finally, the author would like to thank Geoffrey Dorne, graphic designer, for creating a unique font, "Pucci's ideograms", representing the ideograms designed by their inventor in the 1930s.

## 9. Appendices/Supplementary Materials

### Appendix A. Pucci's letters

To the best of our knowledge, Federico Pucci authored four registered letters of considerable historical importance: two addressed to U.S. President Harry S. Truman, and two to the Italian National Research Council (CNR). These letters constitute rare documentary evidence of Pucci's attempt to secure recognition and support for his invention at the very moment when electronic computing was beginning to be publicly associated with automatic translation.

1. Pucci's first letter to President Truman, sent around May 10, 1949, explicitly sought financial backing for the development of his invention.
2. His second letter, dispatched on June 13 and recorded as received on July 25, 1949, appears to have been written primarily to assert the priority of his method, anticipating potential disputes of originality.

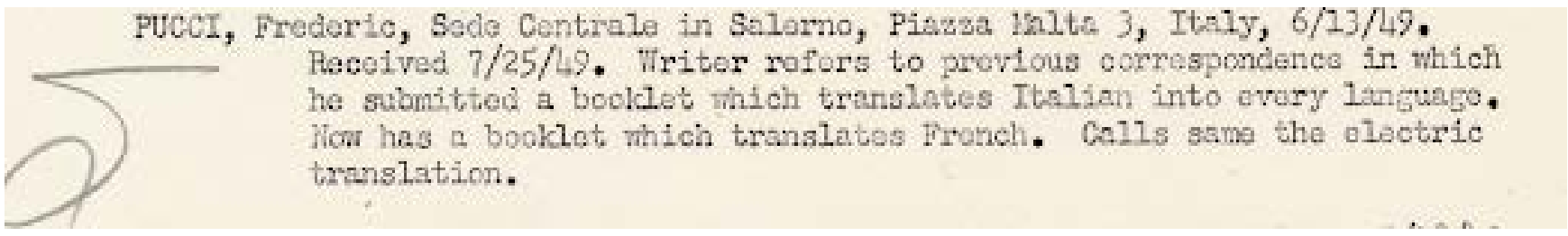
PUCCI, Frederic, Sede Centrale in Salerno, Piazza Malta 3, Italy, 6/13/49. Received 7/25/49. Writer refers to previous correspondence in which he submitted a booklet which translates Italian into every language. Now has a booklet which translates French. Calls same the electric translation.

**Figure A.1**
Official confirmation from the Truman Presidential Library attesting that correspondence was indeed received from one "Frederic [sic] Pucci."

**Transcription**:
*Pucci, Frederic, Sede centrale in Salerno, Piazza Malta 3, Italy, 6/13/49. Received 7/25/49. Writer refers to previous correspondence in which he submitted a booklet which translates Italian into every language. Now has a booklet which translates French. Calls same the electric translation.*

3. On July 10, 1949, Pucci sent a registered letter to the CNR with the same aim of establishing anteriority. This intervention followed his reading in Italian newspaper (unidentified) of a United Press dispatch dated approximately May 31, 1949, which circulated in numerous American and international outlets in early June. The dispatch announced the construction at the University of California, Los Angeles, of an "electric brain" developed by the Bureau of Standards and said to be capable of translating foreign languages. Pucci reacted promptly, positioning his own system against these emerging computational approaches. The CNR responded negatively ten days later, on July 20, 1949.
4. A further letter to the CNR, dated October 17, 1950, reiterated Pucci's claim of anteriority, but the reply of October 27, 1950, once again rejected his petition.

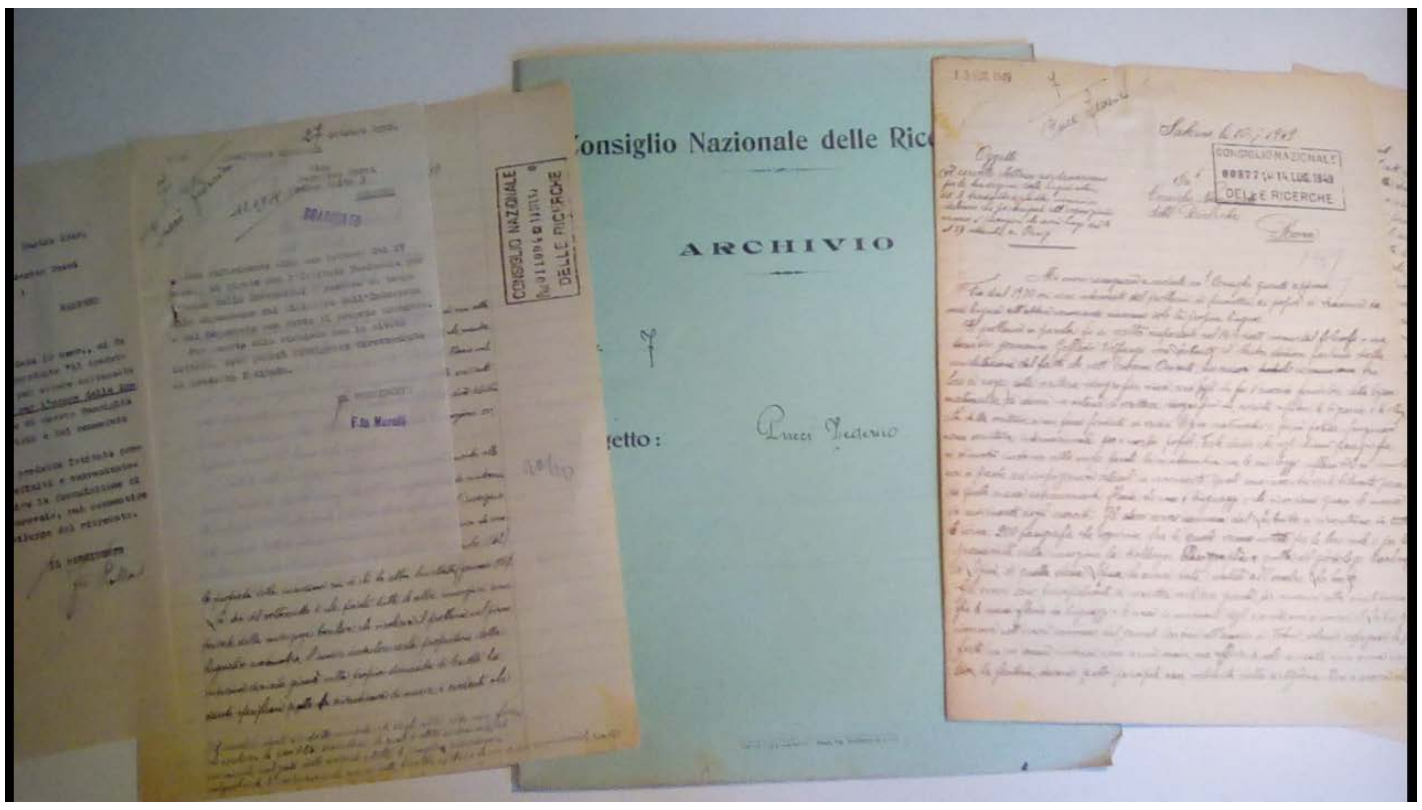

**Figure A.2**
Correspondence between Federico Pucci and the CNR is kept within the Central State Archive of Italy

## Appendix B. Cover of Pucci's 1931 book

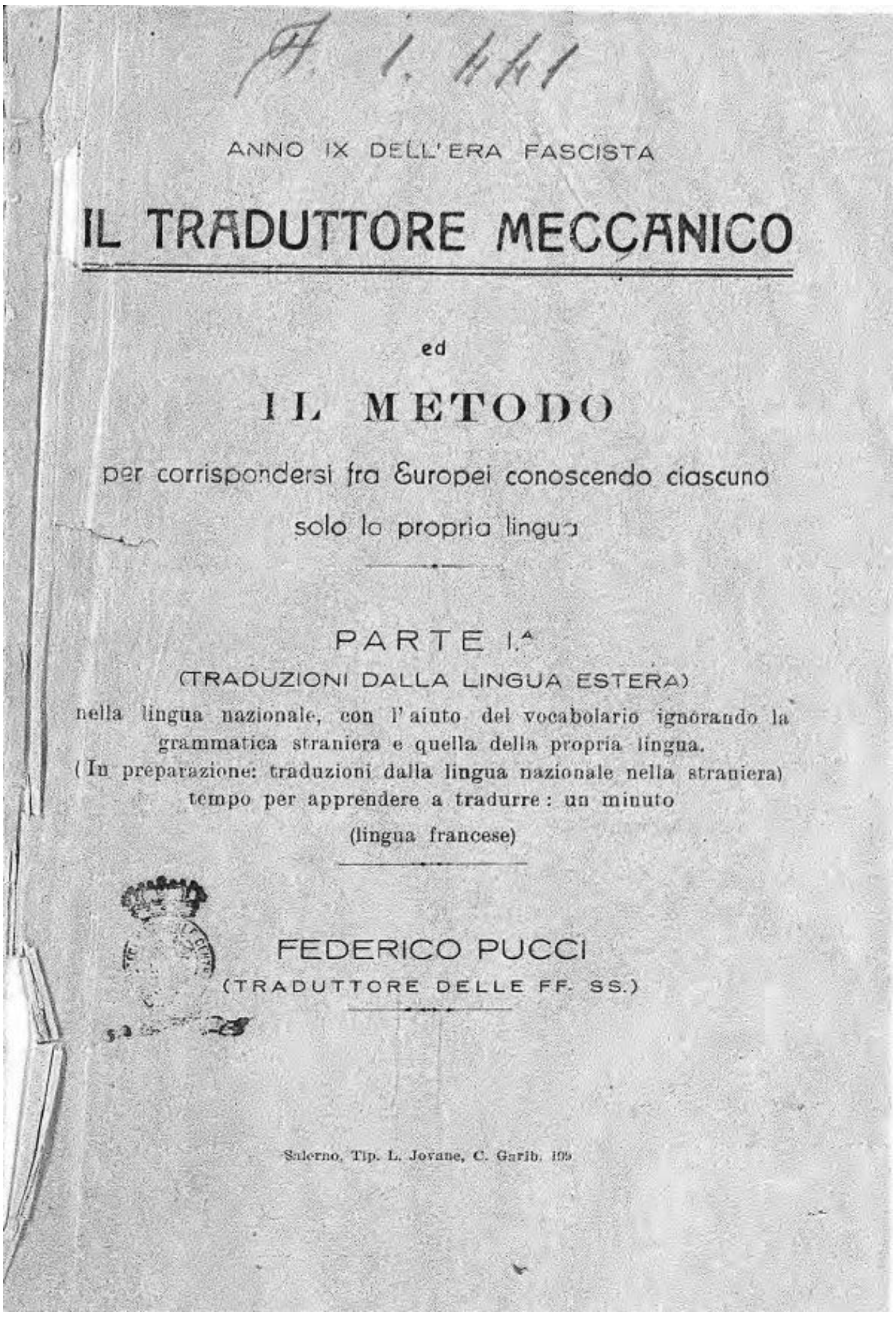
ANNO IX DELL'ERA FASCISTA

IL TRADUTTORE MECCANICO

ed

IL METODO

per corrispondersi fra Europei conoscendo ciascuno solo la propria lingua

PARTE I.ª

(TRADUZIONI DALLA LINGUA ESTERA)

nella lingua nazionale, con l'aiuto del vocabolario ignorando la grammatica straniera e quella della propria lingua.

(In preparazione: traduzioni dalla lingua nazionale nella straniera)

tempo per apprendere a tradurre: un minuto

(lingua francese)

FEDERICO PUCCI

(TRADUTTORE DELLE FF. SS.)

Salerno, Tip. L. Jovane, C. Garib. 199

**Figure B.1**

Pucci's book is kept in the National Central Library of Florence

Author: Pucci, Federico
Title: *Il traduttore meccanico ed il metodo per corrispondersi fra europei conoscendo Ciascuno solo la propria Lingua : Parte I*
Translation: *The Mechanical Translator and the Method for Europeans to Correspond Knowing Only Their Own Language: Part I*
Publication: Salerno, Tip. L. Jovane, 1931
Physical Description: 8vo, 68 pages
BNI Number: BN 1932 1107
Classification: 1 vol. | V.CON F.1.441 | Rare or antique monograph | Available
https://opac.bncf.firenze.sbn.it/Record/CUB0526966

For those who would like to know who Federico Pucci really was: Oriana de Majo, Emanuele Farruggia. 2019. *Federico Pucci censored by Wikipedia*.
https://pucci-in-english.blogspot.com/2025/07/federico-pucci-censored-by-wikipedia.html

## Appendix C. Mentions of Federico Pucci by John Hutchins

As the two brief mentions in early writings by John Hutchins have vanished from the web, I publish the screenshots of original references, which both date back to 1997:

**File MTS-1997.pdf**

[Expanded version of paper given at MT Summit VI, 1997]

**FIRST STEPS IN MECHANICAL TRANSLATION**

John Hutchins
(University of East Anglia, Norwich, UK)

**Abstract**
Although the first ideas for mechanical translation were made in the seventeenth century, it was not until this century that means became available for realisation with the appearance of the electronic computer in the mid 1940s. Fifty years ago, in March 1947 Warren Weaver wrote to Norbert Wiener and met Andrew Booth, mentioning to both the use of computers for translation. The possibilities were investigated during the next seven years, until in January 1954 the first prototype program was demonstrated. This article is a chronicle of these early years of mechanised translation.

**1. The beginnings to 1947** **--> p. 6**

**6. Contemporary proposals**
While we can credit Weaver with the most explicit proposals about MT during the early years of the computer, it is quite clear that the idea of mechanised translation was an idea that had come to others at this time.
There are a number of instances where proposals for translation by computer were put forward independently. In 1946, Duncan Harkin had been inspired by studies of word counts to think of a system based on the most frequent vocabulary, and in collaboration with an electronics engineer, planned the outline of a machine for mechanical translation. In September 1947 Alan Turing wrote a report for the National Physical Laboratory about his plans for constructing a computer, in which he mentioned MT as a possible demonstration of the 'intelligence' of computers. In August 1949, the *New York Times* reported from Salerno that an Italian named Federico Pucci, had invented a machine to translate, saying that it would be exhibited at a Paris Fair; but no more was to be heard of it. In February 1950 Charles C.Holt of Chicago revealed in a letter to the Rockefeller Foundation that he had thought of using cathode ray tubes as a storage device for a mechanical dictionary – in some respects, it was similar in concept to the mechanism put forward by Troyanskii in the 1930s.

**File MTJ-1997-corr.pdf**

[Corrected version (2005) of paper in: *Machine Translation*, vol.12 no.3, 1997, p.195-252]

**From first conception to first demonstration: the nascent years of machine translation, 1947-1954 A chronology**

*John Hutchins*
(http://ourworld.compuserve.com/homepages/WJHutchins)

**File MTJ-1997.pdf**

[From: *Machine Translation*, vol.12 no.3, 1997, p.195-252]

**From first conception to first demonstration: the nascent years of machine translation, 1947-1954 A chronology**

*John Hutchins*
(University of East Anglia, Norwich NR4 7TJ, England)

**Abstract.** The early history of applying electronic computers to the task of translating natural languages is chronicled, from the first suggestions by Warren Weaver in March 1947 to the first demonstration of a working, if limited, program in January 1954.

**Keywords.** Machine translation, history.

**1. The beginnings in March 1947** **--> p. 12**

Although no further details are provided it is evident that the proposal would have done no more than give a word for word translation of a simple text, with the most significant words not being translated at all since the set of 500 would be function words and common vocabulary. The proposal was less ambitious than Troyanskii's (Hutchins 1993), and whether Harkin intended a computer implementation is uncertain.
The second consists of an intriguing announcement from Italy just three months after the newspaper report of SWAC. On 26 August 1949, the *New York Times* reported (page 9) from Salerno:

> Federico Pucci announced today that he had invented a machine that could translate copy from any language into any other language. He said that the machine was electrically operated, but refused to disclose details. He said that he would enter it in the Paris International Fair of Inventions next month.

It is uncertain whether Pucci had any knowledge of Huskey's proposals, and it seems most unlikely he knew about Weaver's memorandum or the British experiments. In any event, there is no trace of any demonstration at the Paris fair; and nothing more is known about Pucci.
It is not known how many others at this time had similar ideas about translating machines. The new electronic computers had caught the imagination of many people. Reports on the 'electronic brains' – the term regularly used by journalists – appeared almost daily in national newspapers throughout the world. Translation was then, and often still is, regarded by those unfamiliar with its difficulties as essentially a question of finding equivalent words in another language. To use a computer in such a task seemed trivial.

**8. The Unesco conference**

**Figure C.1**
Screenshots of original references to Federico Pucci by John Hutchins

*The first document (MTS-1997.pdf) is entitled:*
*[Expanded version of paper given at MT Summit VI, 1997]*
*First Steps in Mechanical Translation*
*John Hutchins (University of East Anglia, Norwich, UK)*

*The second document (MTJ-1997.pdf) bears the title:*
*[From: Machine Translation, vol. 12, no. 3, 1997, pp. 195–252]*
*From First Conception to First Demonstration: The Nascent Years of Machine Translation, 1947–1954. A Chronology*
*John Hutchins (University of East Anglia, Norwich NR4 7TJ, England)*

*The third document (MTJ-1997-corr.pdf) is a corrected version of the second, issued in 2005, which reproduces verbatim the reference to Pucci:*
*[Corrected version (2005) of paper in: Machine Translation, vol. 12, no. 3, 1997, pp. 195–252]*
*From First Conception to First Demonstration: The Nascent Years of Machine Translation, 1947–1954. A Chronology*
*John Hutchins*

The Machine Translation Summit VI held at the Catamaran Resort Hotel in San Diego, California, from October 19 through November 1, 1997, was a gala affair celebrating 50 years of "mechanical translation," opening with the panel "*The Pioneers of Machine Translation.*" In fact, Hutchins placed the beginnings of MT in 1947, 16 (sixteen) years after the publication of Pucci's book.

## Appendix D. The Concepts of "*Plain Language*" and "*Good Enough*"

In modern machine translation post-editing (MTPE) workflows, plain language in the source text improves machine translation output, while "good enough" post-editing enables faster and more cost-effective delivery by avoiding unnecessary over-editing when only functional accuracy is required (e.g., for internal use, gisting, or content monitoring). Together, these complementary concepts support a lean MTPE workflow that is both economical and fit for purpose:

- Plain language functions as an upstream enabler, enhancing input quality (reducing errors) and thereby improving MT accuracy and simplifying the post-editing process.
- Good enough operates as a downstream filter, lowering costs and calibrating the level of editorial effort to the client's actual needs.

Pucci had already articulated these two principles in 1931, though in different terms: (1) simplification of the input language, and (2) simplification of the output text.

1. Simplification of the input language

In his treatment of Dante's *La Vita Nuova*, Pucci proposed simplifying rare or complex expressions into more accessible equivalents before translation. The result is a more transparent input sentence: *appariva vestita di un colore molto nobile, cinta e adorna come conveniva alla sua giovanissima età* (original one: *Apparve vestita di nobilissimo colore, cinta ed ornata alla guisa che alla sua giovanissima età si conveniva)* — a structurally simplified form that facilitates translation:

- *Apparve* is the past simple of *apparire*: Rule #5, which advises replacing uncommon expressions with simpler forms, could very well apply today, at a time when this tense is no longer used so much.
- *Nobilissimo* becomes *molto nobile* under Rule #4, replacing diminutives and augmentatives with straightforward modifiers.
- The phrase *alla guisa che alla sua giovanissima età si conveniva* is simplified by substituting *alla guisa* with *come* and making implicit elements explicit, or the contrary, in line with Rule #3, which recommends expressing propositions in direct constructions

Upstream, simplification involves expressing sentences in direct form, resolving ellipses, and reducing messages to their minimal semantic units. This is, in effect, a precursor to the modern principle of plain language: by reducing linguistic complexity, texts become easier to translate, process, and understand.

2. Simplification of the output text

On the output side, Pucci explicitly prioritized comprehension over stylistic finesse, stressing that "*the translation is quite correct, and... anyway it's not about having a perfect translation, it's just about understanding.*"

Downstream, his model was a "post-editing process" in which the target-language recipient reconstructs the simplified text with the help of standardized ideograms and international keys. The recipient, as a native speaker, reorganizes "machine-generated words" into conventional word order, producing coherent sentences and easily retrieving equivalents for unfamiliar terms. Pucci thus prioritized functional comprehensibility over stylistic perfection.

This system favored accessibility and functional communication over absolute precision. Thus, Pucci's "translation machine," though never built, already combined two principles that underpin current usage: simplification of input (plain language) and production of output that is adequate rather than perfect (good enough). Together, they form a framework balancing accuracy, usability, and efficiency, and anticipated two paradigms that today define efficient MTPE practice.

## Appendix E. Other Pucci's books

As far as we know, Federico Pucci authored ten works dedicated to his invention over a period spanning from 1931 to 1960. Apart the two books previously discussed, the other eight are as follows:

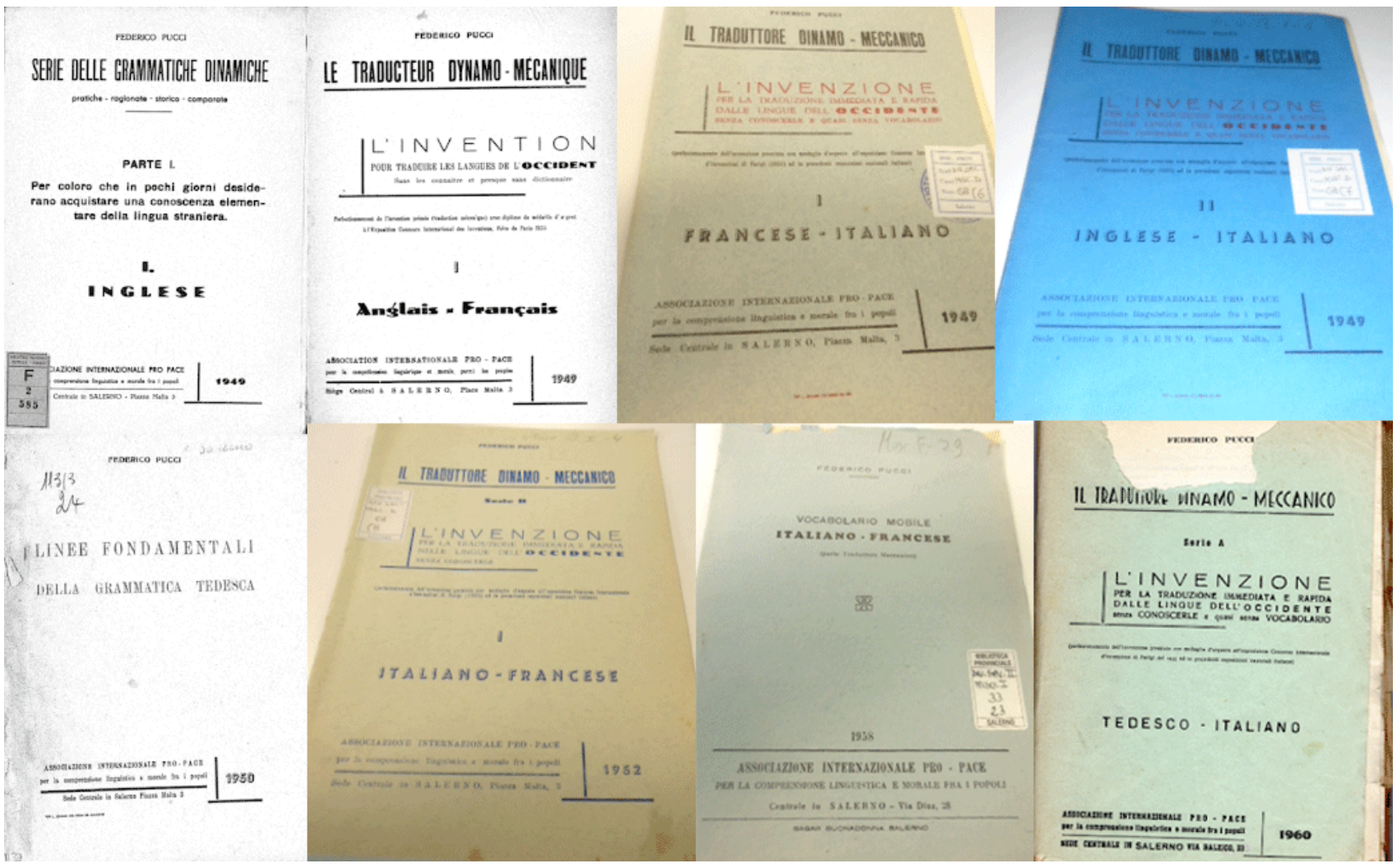

**Figure E.1**
Eight books out of the 10 written by Federico Pucci on his invention

1. **Florence, 1931 (See Appendix B)**
2. Florence, 1949: Series of dynamic, practical, reasoned, historical-comparative grammars: Part I. For those who wish to acquire, in just a few days, an elementary knowledge of the foreign language. Fascicle I. English
3. Salerno, March 1949: The dynamo-mechanical translator: Series A. The invention for immediate and rapid translation of Western languages without knowing them and almost without a dictionary. Fascicle 1. French–Italian
4. Salerno, May 1949: The dynamo-mechanical translator: Series A. The invention for immediate and rapid translation of Western languages without knowing them and almost without a dictionary. Fascicle 2. English–Italian
5. Florence, June 1949 (published in French): Le traducteur dynamo-mecanique: L'invention pour traduire les langues de l'Occident sans les connaître et presque sans dictionnaire. Op. I: anglais–français (sic)
6. Florence, 1950: Dynamic grammar of the German language: Fundamental outlines
7. **Salerno, Florence, 1950 (See Appendix F)**
8. Salerno, 1952: The dynamo-mechanical translator: Series B. The invention for immediate and rapid translation of Western languages without knowing them and almost without a dictionary. Fascicle 1. Italian–French
9. Salerno, 1958: Mobile Italian–French vocabulary (mechanical translator section)
10. Anna Pucci, 1960: The dynamo-mechanical translator: Series A. The invention for immediate and rapid translation of Western languages without knowing them and almost without a dictionary. German–Italian

A copy of this last work, which was absent from all libraries consulted to date (including Salerno and Florence), was most kindly provided to me by Mr. Federico Pucci's daughter, Anna, who also showed me numerous grammars annotated in her father's hand, covering an extraordinary range of languages: from Albanian to Catalan, from Armenian to Somali, including Hebrew and Aramaic, among others. The ten books published by Pucci cover only four languages: Italian, French, English and German.

**Appendix F. The Pucci's "book-machine"**

We have a record of two copies of the book-machine, with this cover:

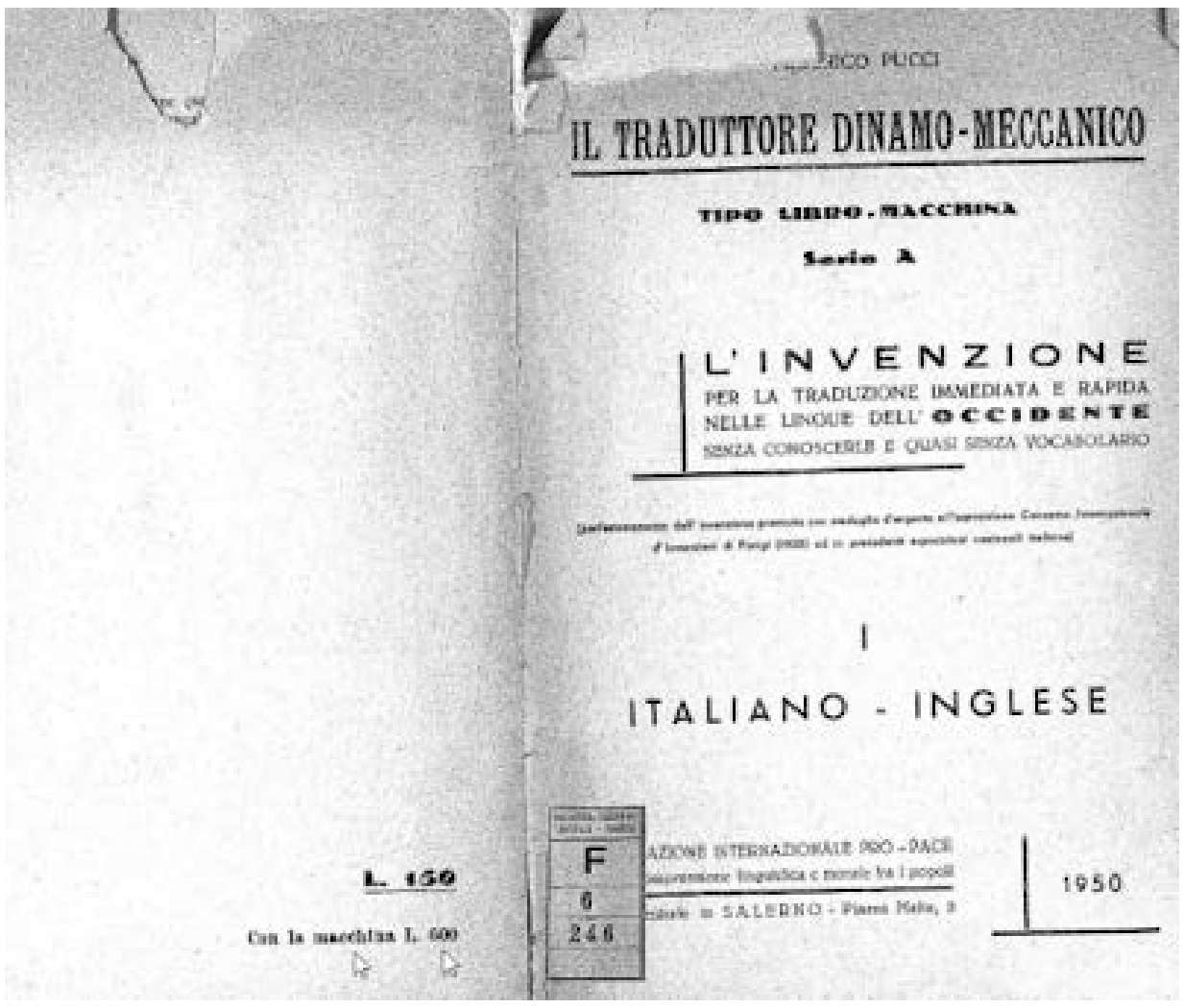

**Figure F.1**

Pucci's book is kept 1) in the National Central Library of Florence, and 2) in the Provincial Library of Salerno

1) Author: Pucci, Federico
Title: *Il traduttore dinamo-meccanico : Tipo libro macchina. Serie a. L'invenzione per la traduzione immediata e rapida nelle lingue dell'Occidente senza conoscerle e quasi senza vocabolario. [fasc. ] I. Italiano-Inglese*
Translation: *The Dynamo-Mechanical Translator: Book Machine Type. Series A. The invention for immediate and rapid translation into Western languages without knowing them and almost without a dictionary. [Fascicle] I. Italian-English*
Publication: Salerno: Ed. Ass. Internaz. Pro-Pace, 1950
Physical Description: 8vo, (64) pages
BNI Number: BN 1951 1794
Classification: 1 vol. | V.CON F.6.246.1 | Document for consultation only | Available
https://opac.bncf.firenze.sbn.it/Record/CUB0526965

2) [M] Pucci, Federico - Il *traduttore dinamo-meccanico : Tipo libro macchina. Serie a. L'invenzione per la traduzione immediata e rapida nelle lingue dell'Occidente senza conoscerle e quasi senza vocabolario. [fasc. ] 1. Italiano-Inglese / Federico Pucci. - Salerno : Ed. Ass. Internaz. Pro-Pace , 1950 [Document Code: CAM0053518]

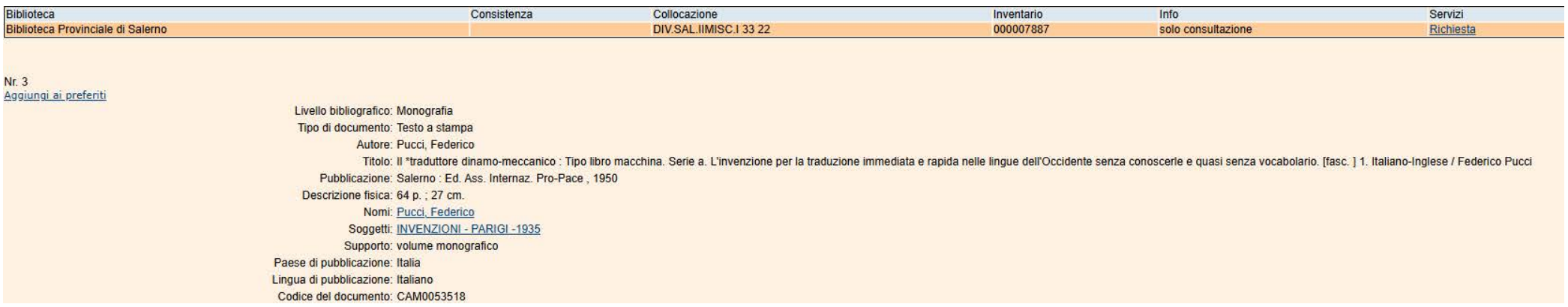

| Biblioteca | Consistenza | Collocazione | Inventario | Info | Servizi |
|---|---|---|---|---|---|
| Biblioteca Provinciale di Salerno | | DIV.SAL.IIMISC.I 33 22 | 000007887 | solo consultazione | Richiesta |

Nr. 3
Aggiungi ai preferiti

Livello bibliografico: Monografia
Tipo di documento: Testo a stampa
Autore: Pucci, Federico
Titolo: Il *traduttore dinamo-meccanico : Tipo libro macchina. Serie a. L'invenzione per la traduzione immediata e rapida nelle lingue dell'Occidente senza conoscerle e quasi senza vocabolario. [fasc. ] 1. Italiano-Inglese / Federico Pucci
Pubblicazione: Salerno : Ed. Ass. Internaz. Pro-Pace , 1950
Descrizione fisica: 64 p. ; 27 cm.
Nomi: Pucci, Federico
Soggetti: INVENZIONI - PARIGI -1935
Supporto: volume monografico
Paese di pubblicazione: Italia
Lingua di pubblicazione: Italiano
Codice del documento: CAM0053518

**Figure F.2**

Pucci's monograph kept in the Provincial Library of Salerno:
Subjects: INVENTIONS - PARIS - 1935

As we can see on Figure C.1, Pucci imagined his translator as an affordable device (priced at 450 lire in his vision, that is three times the price of the book without the machine, less than 12 USD today), and envisioned various forms of the machine, including mechanical, electrical, phono-electrical, photo-electrical, and tele-electrical versions, as well as a portable “electro-mechanical interpreter” (*Interprète électro-mécanique portable*). These descriptions appear in this book.

However, these were only conceptual models, often accompanied by blueprints or theoretical diagrams rather than functional devices. For instance, a small cardboard model annotated by Pucci himself was found in the Salerno Provincial Library and in the National Central Library of Florence, described as a “book-machine” for immediate translation without a dictionary, but it was not a working device:

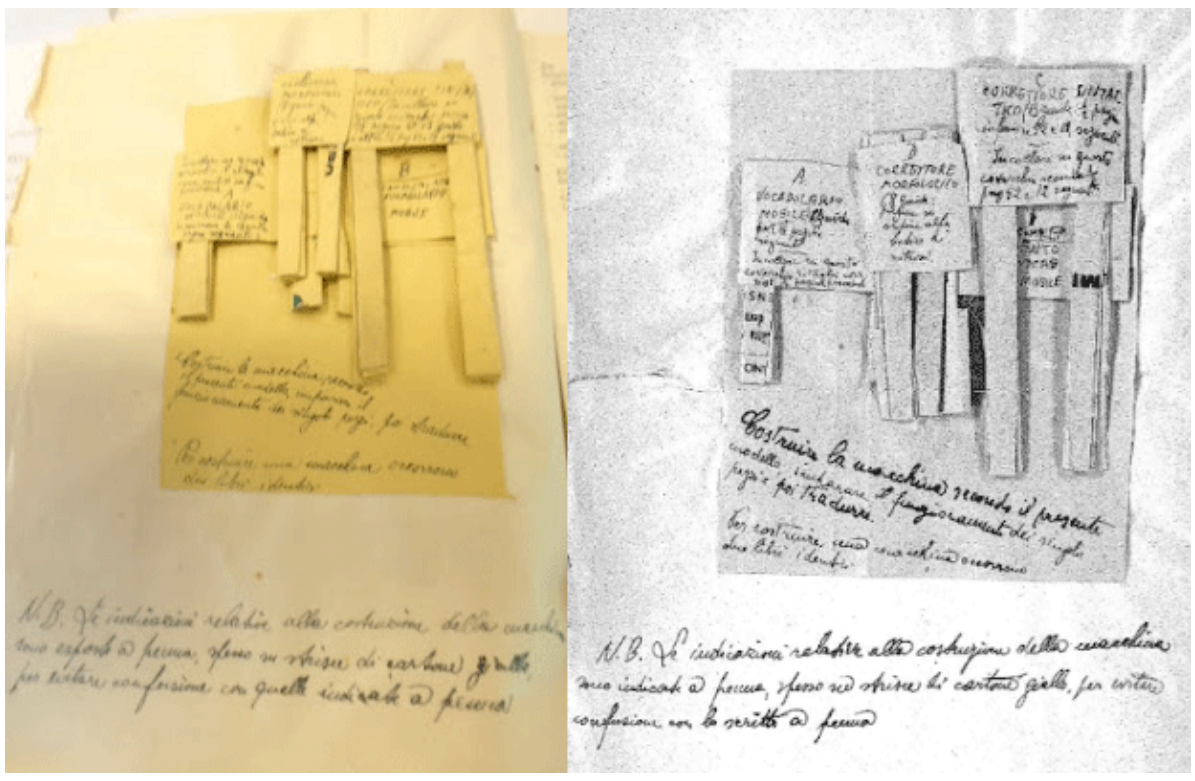

**Figure F.3**

Handmade model with handwritten annotations by Pucci, kept 1) in the Provincial Library of Salerno (on the left), and 2) in the National Central Library of Florence (on the right)

Pucci’s ideograms and keys were part of this modular “book-machine” system, which included:

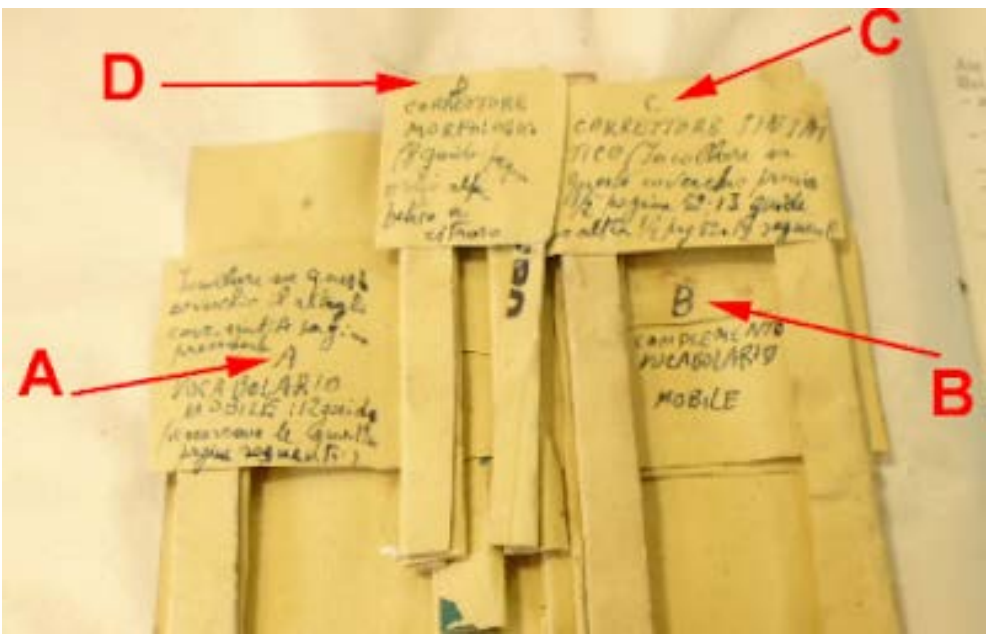

**Figure F.4**

Handwritten description of the modules by Pucci

- Mobile Vocabulary (A): A dictionary of base words.
- Supplement to the Mobile Vocabulary (B): Additional terms or variations.
- Syntactic Corrector (C): Rules for arranging words in the target language’s syntax.
- Morphological Corrector (D): Adjustments for inflections (e.g., conjugations, declensions).

The print run of this work remains unknown; however, given that each mock-up was meticulously produced and annotated manually by Pucci, one can readily infer the considerable amount of time such an undertaking must have required.

## Appendix G. Quantitative comparison between removals and additions

Comparison between removals and additions, where removals designate linguistic units (a word or short sequence) present in version A but absent or modified in version B, while additions refer to units appearing in version B that are not found in version A.

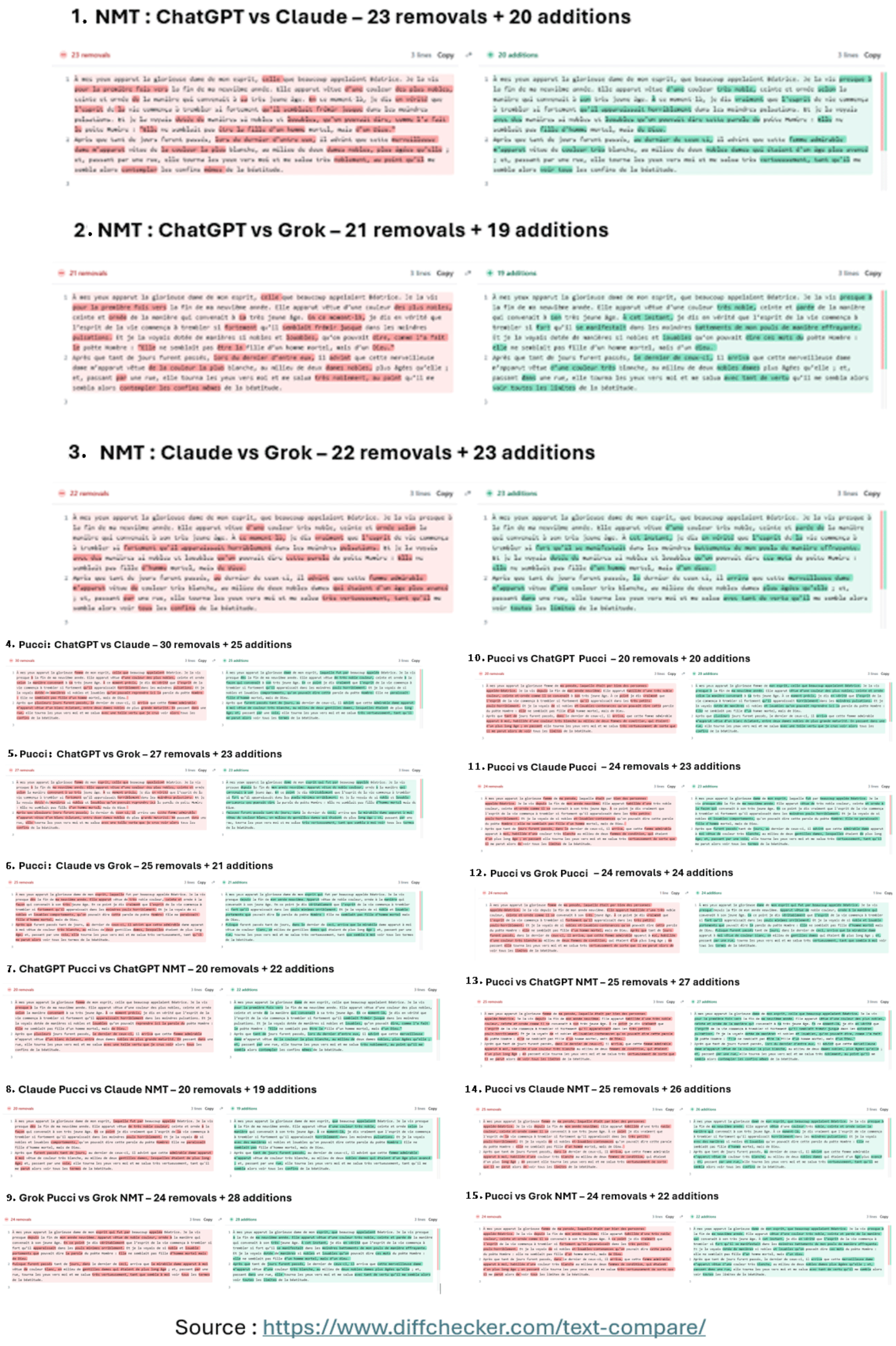

**Figure G.1**
Source of comparison: https://www.diffchecker.com/

## Appendix H. Excerpt from Dante's *Vita Nuova* chosen by Pucci

https://digitaldante.columbia.edu/text/library/la-vita-nuova/#cap2

**II**

[I] Nove fiate già appresso lo mio nascimento era tornato lo cielo de la luce quasi a uno medesimo punto, quanto a la sua propria girazione, quando a li miei occhi apparve prima la gloriosa donna de la mia mente, la quale fu chiamata da molti Beatrice, li quali non sapeano che si chiamare. Ella era in questa vita già stata tanto, che ne lo suo tempo lo cielo stellato era mosso verso la parte d'oriente de le dodici parti l'una d'un grado, sì che quasi dal principio del suo anno nono apparve a me, ed io la vidi quasi da la fine del mio nono. Apparve vestita di nobilissimo colore, umile ed onesto, sanguigno, cinta e ornata a la guisa che a la sua giovanissima etade si convenia. In quello punto dico veracemente che lo spirito de la vita, lo quale dimora ne la secretissima camera de lo cuore, cominciò a tremare sì fortemente che apparia ne li mènimi polsi orribilmente; e tremando, disse queste parole: «Ecce deus fortior me, qui veniens dominabitur mihi». In quello punto lo spirito animale, lo quale dimora ne l'alta camera ne la quale tutti li spiriti sensitivi portano le loro percezioni, si cominciò a maravigliare molto, e parlando spezialmente a li spiriti del viso, sì disse queste parole: «Apparuit iam beatitudo vestra». In quello punto lo spirito naturale, lo quale dimora in quella parte ove si ministra lo nutrimento nostro, cominciò a piangere, e piangendo, disse queste parole: «Heu miser, quia frequenter impeditus ero deinceps!». D'allora innanzi dico che Amore segnoreggiò la mia anima, la quale fu sì tosto a lui disponsata, e cominciò a prendere sopra me tanta sicurtade e tanta signoria per la vertù che li dava la mia imaginazione, che me convenia fare tutti li suoi piaceri compiutamente. Elli mi comandava molte volte che io cercasse per vedere questa angiola giovanissima; onde io ne la mia puerizia molte volte l'andai cercando, e vedèala di sì nobili e laudabili portamenti, che certo di lei si potea dire quella parola del poeta Omero: *Ella non parea figliuola d'uomo mortale, ma di Deo*. E avegna che la sua imagine, la quale continuamente meco stava, fosse baldanza d'Amore a segnoreggiare me, tuttavia era di sì nobilissima vertù, che nulla volta sofferse che Amore mi reggesse sanza lo fedele consiglio de la ragione in quelle cose là ove cotale consiglio fosse utile a udire. E però che soprastare a le passioni e atti di tanta gioventudine pare alcuno parlare fabuloso, mi partirò da esse; e trapassando molte cose, le quali si potrebbero trarre de l'esemplo onde nascono queste, verrò a quelle parole le quali sono scritte ne la mia memoria sotto maggiori paragrafi.

https://digitaldante.columbia.edu/text/library/la-vita-nuova/#cap3

**III**

[II] Poi che furono passati tanti die, che appunto erano compiuti li nove anni appresso l'apparimento soprascritto di questa gentilissima, ne l'ultimo di questi die avvenne che questa mirabile donna apparve a me vestita di colore bianchissimo, in mezzo a due gentili donne, le quali erano di più lunga etade; e passando per una via, volse li occhi verso quella parte ov'io era molto pauroso, e per la sua ineffabile cortesia, la quale è oggi meritata nel grande secolo, mi saluto e molto virtuosamente, tanto che me parve allora vedere tutti li termini de la beatitudine.

**Figure H.1**
Pieces of sentences assembled by Pucci to construct his excerpt.
Source: https://digitaldante.columbia.edu/text/library/la-vita-nuova/

**Pucci's excerpt from Dante:**
*Ai miei occhi apparve la gloriosa donna della mia mente, la quale fu da molti chiamata Beatrice.*
*Io la vidi quasi dalla fine del mio anno nono. Apparve vestita di nobilissimo colore, cinta ed ornata alla guisa che alla sua giovanissima età si conveniva. In quel punto dico veramente che lo spirito della vita cominciò a tremar si fortemente che apparia nei menomi polsi orribilmente.*
*E vedeala di si nobili e laudabili portamenti, che si potea dire quella parola del poeta Omero:* Ella non parea figliuola d'uomo mortale, ma di Dio.
*Poi che furono passati tanti dì,*
*nell'ultimo di questi, avvenne che questa mirabile donna apparve a me vestita di colore bianchissimo, in mezzo di due gentili donne, le quali erano di più lunga età; e, passando per una via, volse gli occhi verso me*
*e mi salutò molto virtuosamente, tanto che mi parve allora vedere tutti i termini della beatitudine.*

## Appendix I. Candidate's translations before tokenization

### 1. Maxime Durand-Fardel (1898)

*Quand apparut à mes yeux pour la première fois la glorieuse dame de mes pensées, que beaucoup nommèrent Béatrice. De sorte qu'elle était au commencement de sa neuvième année, quand elle m'apparut, et moi à la fin de la mienne. Je la vis vêtue … d'une façon simple et modeste, et parée comme il convenait à un âge aussi tendre. À ce moment, je puis dire véritablement que le principe de la vie … se mit à trembler si fortement en moi que je le sentis battre dans toutes les parties de mon corps d'une façon terrible. Et je lui voyais une apparence si noble et si belle que certes on pouvait lui appliquer cette parole d'Homère. « Elle paraissait non la fille d'un homme mais celle d'un Dieu. »*
*Après que furent passées neuf années … le dernier jour, je la rencontrai vêtue de blanc, entre deux dames plus âgées. Comme elle passait dans une rue, elle jeta les yeux du côté où je me trouvais … et, avec une courtoisie infinie, … , elle me salua si gracieusement qu'il me sembla avoir atteint l'extrémité de la Béatitude.*

### 2. Godefroy (1901)

*Quand à mes yeux apparut la glorieuse dame de mon esprit qui, par beaucoup de personnes … fut appelée Béatrice, et je l'ai vue presque à la fin de ma neuvième année. Elle m'apparut vêtue d'une très noble couleur, … rose et ornée comme il convenait à un âge aussi délicieux d'une aussi grande jeunesse, mais de telle sorte que je commençais à frissonner si fortement que, dans ses moindres mouvements, elle m'apparut terriblement… C'est ainsi que j'ai pu voir des maintiens si nobles et si admirables, que l'on pouvait pour elle prononcer sans crainte la parole du poète Homère : « Elle paraissait non la fille d'un homme, mais celle d'un Dieu. »*
*Une fois passés tant de jours, … ce fut le dernier jour que cette Dame m'apparut, vêtue de couleur blanche, entre deux autres nobles dames qui étaient d'un bien plus grand âge. Tout en passant par un chemin, … elle me salua si gracieusement qu'il me sembla surpasser tous les confins de la Béatitude.*

### 3. Cochin (1905)

*Lorsqu'à mes yeux apparut premièrement la glorieuse Dame de mon âme, laquelle fut nommée Béatrice par bien des gens. Et moi je la vis à peu près à la fin de ma neuvième [année]. Et elle m'apparut vêtue de très noble couleur, … ceinte et parée en la manière qui était convenable à son très jeune âge. En ce moment, je dis véritablement que l'Esprit de la Vie, … commença à trembler si fortement, qu'il se faisait sentir en les plus petites veines terriblement; et je voyais en elle des façons si nobles et si louables que certes on pouvait dire d'elle cette parole du poète Homère : « Elle ne paraissait pas fille d'un homme mortel, mais de Dieu. »*
*Après que se furent passés assez de jours … en le dernier de ces jours, il advint que cette admirable Dame m'apparut, vêtue de couleur très blanche, au milieu de deux, gentilles dames, qui étaient d'âge plus avancé ; et, passant par une rue, elle tourna les yeux du côté où je me trouvais, … elle me salua d'une telle grâce qu'il me sembla voir alors toutes les limites de la béatitude.*

### 4. ChatGPT 5 (2025)

*À mes yeux apparut la glorieuse dame de mon esprit, que beaucoup nommaient Béatrice. Je la vis presque à la fin de ma neuvième année. Elle apparut vêtue d'une couleur très noble, ceinte et ornée de la manière qui convenait à son très jeune âge. En ce moment, je dis en vérité que l'esprit de la vie commença à trembler si fortement qu'il se manifestait jusque dans les moindres pulsations, horriblement. Et je la voyais douée de si nobles et louables manières, qu'on pouvait répéter ce mot du poète Homère : Elle ne paraissait pas fille d'un homme mortel, mais de Dieu.*
*Après que tant de jours se furent écoulés, le dernier d'entre eux, il advint que cette admirable dame m'apparut vêtue d'une blancheur éclatante, au milieu de deux nobles dames plus âgées qu'elle ; et, passant par une rue, elle tourna les yeux vers moi et me salua avec une grande vertu, de sorte qu'il me sembla alors contempler toutes les limites de la béatitude.*